\documentclass[10pt,twocolumn,letterpaper]{article}

\usepackage[pagenumbers]{iccv} % To force page numbers, e.g. for an arXiv version

\usepackage[dvipsnames]{xcolor}

\def\A{{\bf A}}

\def\B{{\bf B}}

\def\K{{\bf K}}

\def\P{{\bf P}}

\def\X{{\bf X}}
\def\Y{{\bf Y}}
\def\Q{{\bf Q}}

\def\U{{\bf U}}

\def\V{{\bf V}}

\def\W{{\bf W}}

\def\0{{\bf 0}}
\def\1{{\bf 1}}

\def\SigmaBM{\boldsymbol{\Sigma}}

\def\RB{{\mathbb R}}

\def\Tha{\mbox{\boldmath$\Theta$\unboldmath}}

\usepackage{indentfirst}

\definecolor{iccvblue}{rgb}{0.21,0.49,0.74}
\usepackage[pagebackref,breaklinks,colorlinks,allcolors=iccvblue]{hyperref}

\def\paperID{3733} % *** Enter the Paper ID here
\def\confName{ICCV}
\def\confYear{2025}

\usepackage[utf8]{inputenc} % allow utf-8 input
\usepackage[T1]{fontenc}    % use 8-bit T1 fonts
\usepackage{hyperref}       % hyperlinks

\usepackage{url}            % simple URL typesetting
\usepackage{graphicx}
\usepackage{multirow}
\usepackage{colortbl}
\usepackage{caption, subcaption}
\usepackage{booktabs}  
\usepackage{arydshln}  
\usepackage{xcolor}

\usepackage{amsmath}  
\usepackage{wrapfig}
\usepackage{microtype}
\usepackage{pifont}  

\usepackage{marvosym}  % \Letter

\definecolor{lightgrey}{rgb}{0.85,0.85,0.85}
\definecolor{lightlightgrey}{rgb}{0.9,0.9,0.9}

\definecolor{col1}{RGB}{255, 230, 230}
\definecolor{col2}{rgb}{0.9372, 0.5804, 0.6196}
\definecolor{col3}{RGB}{233, 255, 245}

\title{Attention to the Burstiness in Visual Prompt Tuning!}
\author{
  Yuzhu Wang$^{1}$
  \quad Manni Duan$^{1}$ 
  \quad Shu Kong$^{2,3,\text{\Letter}}$ \\ 
  { 
  $^1$Zhejiang Lab \quad $^2$University of Macau \quad $^3$Institute of Collaborative Innovation}  
  \\
  {code: \url{https://github.com/WangYZ1608/BPT}}
  \vspace{-0mm}
}

\begin{document}

\twocolumn[
{%
\vspace{-3mm}
\maketitle
\vspace{-2em}
% \begin{@twocolumn}
\centering
\includegraphics[width=1.0\linewidth]{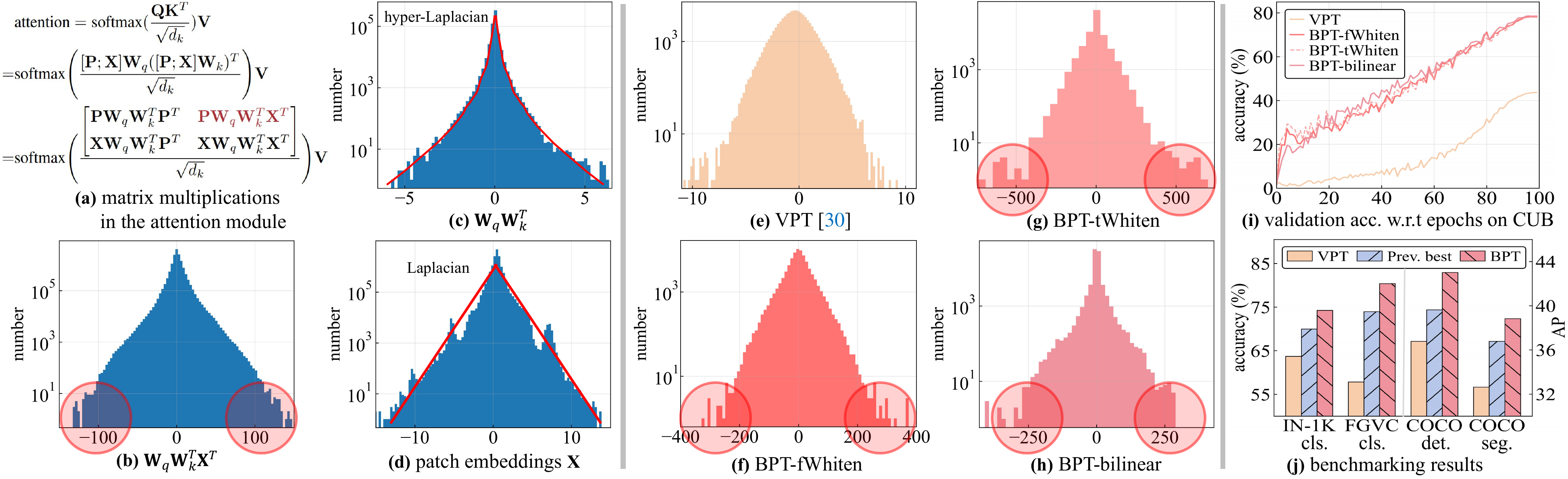} 
\vspace{-7mm}
\captionof{figure}{\small
{\bf Attention to the burstiness in Visual Prompt Tuning (VPT)!}
VPT~\cite{jia2022visual} learns a small number of parameters (so-called prompts) $\P$ in the input space with a Transformer~\cite{dosovitskiy2020image}: $\P$ is multiplied by image patch embeddings $\X$, the query $\W_q$ and key $\W_k$ projectors in the attention module {\bf (a)}.
We uncover \emph{burstiness}~\cite{jegou2009burstiness} in them: 
a small portion of entries in ${\color{Maroon}\W_q \W_k^T\X^T}$ have very large absolute values {\bf (b)}.
Further, we uncover non-Gaussian distributions: 
values of $\W_q\W_k^T$ follow a hyper-Laplacian distribution~\cite{krishnan2009fast} {\bf (c)}, and 
values of patch embedding $\X$ follow a Laplacian distribution {\bf (d)}.
Intuitively, such non-Gaussian distributions pose challenges in learning prompt $\P$.
To facilitate learning, we propose \emph{whitening} \cite{friedman1987exploratory, ranzato2010factored} ${\color{Maroon}\W_q \W_k^T\X^T}$, transforming it to be more Gaussian before learning prompt $\P$.
We derive the whitening matrix $\W$ and fix it to learn prompt $\P$ (dubbed BPT-fWhiten), resulting into the effective prompt as $\tilde\P = \P \W^T$.
Surprisingly, BPT-fWhiten significantly accelerates learning and boosts accuracy {\bf (i)}, \eg, from 42.15\% by VPT to 77.48\% on the CUB-200 benchmark~\cite{wah2011caltech}!
We compare its value distribution with that of $\P$ learned by VPT in {\bf (e)} and {\bf (f)}.
Interestingly, different from non-bursty prompts learned by VPT~\cite{jia2022visual}, 
BPT-fWhiten exhibits burstiness: a small number of entries in $\tilde\P$ have large values ($>$200).
The reason is that its final prompt $\tilde\P=\P\W^T$ is a \emph{bilinear} operation, which is known to produce burstiness~\cite{lin2015bilinear, kong2017low}.
Significant gains achieved by BPT-fWhiten suggest that learning bursty prompts helps in prompt tuning.
As the Transformer is differentiable w.r.t the whitening matrix $\W$, we finetune it too when learning prompt $\P$ (dubbed BPT-tWhiten), which also produces bursty prompt {\bf (g)}.
With the bilinear operation, we finally present low-rank bilinear prompt tuning (dubbed BPT-bilinear),
which learns two compact matrices $\A$ and $\B$ and use their product as the final prompt, \ie,  $\tilde\P=\A\B^T$. 
BPT-bilinear produces bursty prompts {\bf (h)} and significantly accelerates learning and boosts performance, as shown in {\bf (i)} and {\bf (j)}. Refer to Fig.~\ref{fig:framework} for a conceptual demonstration of our BPT methods.
}
\label{fig:splashy-figure}
\vspace{4mm}
}
]

\begin{abstract}
Visual Prompt Tuning (VPT) is a parameter-efficient fune-tuning technique that adapts a pre-trained vision Transformer (ViT) by learning a small set of parameters in the input space, known as prompts. In VPT, we uncover ``burstiness'' in the values arising from the interaction of image patch embeddings, and the key and query projectors within Transformer's self-attention module. Furthermore, the values of patch embeddings and the key and query projectors exhibit Laplacian and hyper-Laplacian distribution, respectively. Intuitively, these non-Gaussian distributions pose challenges for learning prompts. To address this, we propose whitening these data, 
de-correlating them and equalizing their variance towards more Gaussian before learning prompts. We derive the whitening matrix over random image patch embeddings and ViT's key and query projectors, and multiply it with the prompt to be learned in a bilinear manner.
Surprisingly, this method significantly accelerates prompt tuning and boosts accuracy, e.g., $>$25 accuracy points on the CUB dataset; interestingly, it learns ``bursty prompts''.
Extending the bilinear model which is known to introduce burstiness, we present a compact, low-rank version by learning two smaller matrices whose multiplication yields the final prompts. 
We call the proposed methods Bilinear Prompt Tuning (BPT). Extensive experiments across multiple benchmark datasets demonstrate that BPT methods not only outperform various VPT methods but also reduce parameter count and computation overhead. 
\end{abstract}

\section{Introduction}
\label{sec:intro}

Utilizing vision foundation models~\cite{dosovitskiy2020image, chen2021mocov3, he2022masked, he2020momentum} can significantly boost performance on downstream tasks~\cite{li2022exploring} but adapting them is challenging due to their significant amount of parameters~\cite{dehghani2023scaling}. 
Parameter-efficient fune-tuning fine-tuning (PEFT) attempts to learn a small amount of parameters to adapt foundation models to specific downstream tasks~\cite{houlsby2019parameter, liu2022few, he2021towards}. 
Existing PEFT methods learn bias terms~\cite{zhai2019large}, lightweight adapters~\cite{houlsby2019parameter, hu2021lora}, or linear layers~\cite{lian2022scaling}.
Among them, Visual Prompt Tuning (VPT)~\cite{jia2022visual}, or generally prompt tuning beyond vision~\cite{li2021prefixtuning, lester2021power, liu2022ptuning}, shows better performance on diverse tasks~\cite{yoo2023improving, singh2023effectiveness, wang2024revisiting}.

\textbf{Status quo.}
With a Transformer architecture~\cite{dosovitskiy2020image},
VPT learns a small number of parameters, so-called \emph{prompts} $\P \in \RB^{m\times d}$, in the input space~\cite{jia2022visual}.
The learned prompt $\P$ and an input image's $n$ $d$-dimensional patch embeddings (also called image tokens) $\X\in \RB^{n\times d}$ are concatenated (\ie, $[\P;\X] \in \RB^{(n+m)\times d}$) and interact with the query projector $\W_q \in \RB^{d\times d_k}$ and the key projector $\W_k \in \RB^{d\times d_k}$ within the Transformer's self-attention module\footnote{To simplify presentation, we take as an example the first layer where the input is image tokens $\X$ and the learned prompt $\P$.
We also drop the learnable classification token.} (Fig.~\ref{fig:splashy-figure}\textcolor{iccvblue}{a}):
{\small
\begin{align}
& \text{attention} = \text{softmax}(\frac{\Q \K^T}{\sqrt{d_k}})\V 
\label{eq:attention1}
\\
=   & \text{softmax}\Bigg(\frac{ [\P; \X]\W_q ([\P;\X]\W_k)^T}{\sqrt{d_k}}\Bigg)  
        \V
\label{eq:attention2} 
\\ 
= & \text{softmax}\Bigg(\frac{
        \begin{bmatrix}
            {\color{Maroon}\P\W_q \W_k^T\P^T} & {\color{Maroon}\P\W_q \W_k^T\X^T} \\[0.4em]
            {\color{Maroon}\X\W_q \W_k^T\P^T} & \X\W_q \W_k^T\X^T \\
        \end{bmatrix}
        }{\sqrt{d_k}}\Bigg) 
        \V
\label{eq:attention3}
\end{align}}
where $\V = [\P;\X]\W_v \in \RB^{{(n+m)}\times d_v}$ is the  \emph{value} in the attention module.
Follow-up works improve VPT in different aspects.
For example, GateVPT~\cite{yoo2023improving} learns a gate module to automatically determine at which Transformer blocks to learn prompts.
Some works~\cite{tu2023visual, cheng2023e2vpt} append prompts only to query or key tokens in attention layers to reduce training cost.  SPT~\cite{wang2024revisiting} initializes prompts using clusters of image patch tokens, achieving the state-of-the-art on various benchmarks.

\begin{figure}[t]
    \centering
    \includegraphics[width=0.99\linewidth]{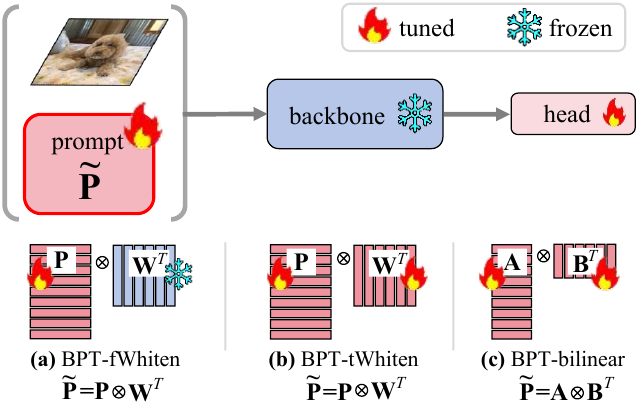}
    \vspace{-1mm}
\caption{\small
    To overcome the challenges due to the burstiness and non-Gaussian distributions (Fig.~\ref{fig:splashy-figure}), we introduce \emph{Bilinear Prompt Tuning (BPT)} with three approaches (\S\ref{sec:method}): 
    {\bf (a)} and {\bf (b)} are based on data whitening, which transforms the non-Gaussian ${\color{Maroon}\W_q \W_k^T\X^T}$ to be more Gaussian through a whitening matrix $\W$. 
    Their difference lies in whether tuning $\W$ when learning prompt $\P$.
    Note that Transformer’s parameters have burstiness and are frozen in prompt tuning;
    {\bf (c)} learns  two compact matrices which are multiplied towards the final prompt $\P$.
    These three approaches share the same bilinear form in prompt learning.
}
\vspace{-2mm}
\label{fig:framework}
\end{figure}

\textbf{Motivation.} 
The significant affects of prompt initialization by SPT~\cite{wang2024revisiting} intrigues us to investigate why learning prompts is difficult.
Therefore, we delve into the distributions of data, a product of prompt $\P$ with query projector, key projector, and image patch embeddings in the self-attention module.
We uncover ``\emph{burstiness}''~\cite{jegou2009burstiness} in them:
a small portion of entries in ${\color{Maroon}\W_q \W_k^T\X^T}$ have very large absolute values  (Fig.~\ref{fig:splashy-figure}\textcolor{iccvblue}{b}).
This is due to the multiplication of several matrices (\ie, multiplying query and key projectors and image tokens) within the attention module, which resembles bilinear operations~\cite{lin2015bilinear, kong2017low}.
Furthermore, we uncover that the values of $\W_q\W_k^T$ follow a hyper-Laplacian distribution~\cite{krishnan2009fast}  (Fig.~\ref{fig:splashy-figure}\textcolor{iccvblue}{c}), and 
the values of patch embedding $\X$ follow a Laplacian distribution  (Fig.~\ref{fig:splashy-figure}\textcolor{iccvblue}{d}).
Intuitively, such bustiness and non-Gaussian distributions present challenges in prompt learning.
We are motivated to address these issues to facilitate prompt tuning.

{\bf Technical insights.}
We adopt data whitening~\cite{friedman1987exploratory, ranzato2010factored} and bilinear model~\cite{lin2015bilinear, kong2017low} facilitate prompt learning.
The former transforms non-Gaussian ${\color{Maroon}\W_q \W_k^T\X^T}$ to be decorrelated and of identity variance, hence more Gaussian, before learning prompt $\P$.
Specifically, we derive the whitening matrix $\W$ and fix it in learning prompt $\P$ (dubbed BPT-fWhiten), resulting into the final \emph{effective} prompt as $\tilde\P=\P\W^T$.
This method significantly accelerates learning (Fig.~\ref{fig:splashy-figure}\textcolor{iccvblue}{i}) and boosts performance (Fig.~\ref{fig:splashy-figure}\textcolor{iccvblue}{j}).
Interestingly, compared with the distribution of values in the learned prompt $\P$ by standard VPT (Fig.~\ref{fig:splashy-figure}\textcolor{iccvblue}{e}),
the final $\tilde \P$ has burstiness (Fig.~\ref{fig:splashy-figure}\textcolor{iccvblue}{f}).
Significant gains achieved by BPT-fWhiten over VPT suggest that learning bursty prompts helps in prompt tuning a Transformer.
The latter, bilinear model, is known to learn parameters with burstiness~\cite{lin2015bilinear, kong2017low}. 
Given the bilinear form of $\tilde\P=\P\W^T$ and its significant performance gains,
we futher propose to learn two compact matrices $\A$ and $\B$, multiplying which leads to the final \emph{low-rank} prompt $\tilde\P=\A\B^T$ (Fig.~\ref{fig:splashy-figure}\textcolor{iccvblue}{h}).
This approach leads to the best results with reduced parameter count and computation overhead.

We call these methods \emph{Bilinear Prompt Tuning (BPT)}, as these methods share the same bilinear format, \ie, the final prompt is the multiplication of two matrices (Fig.~\ref{fig:framework}).
Extensive experiments demonstrate that all our BPT approaches significantly outperform existing VPT methods and other PEFT methods across a range of model scales, dataset sizes,  pre-training objectives, and downstream tasks, achieving state-of-the-art results on various benchmarks (\Cref{tab:sota_ssl,tab:sota_sup,tab:det}).

\textbf{Contributions.} We make three major contributions.
\begin{enumerate}
    \item We uncover \emph{burstiness} and non-Gaussian distributions in the query projector, key projector, and image tokens, which interact with the prompt to be learned. These help explain the difficulty of prompt learning.
    \item We present a multitude of novel solutions named Bilinear Prompt Tuning by relating the burstiness and non-Gaussian distributions to data whitening and bilinear model, both of which involve two-matrix multiplications.
    \item We conduct extensive experiments to validate our BPT methods, demonstrating that they significantly outperform VPT and its variants, as well as other parameter-efficient fine-tuning approaches on various benchmarks.
    
\end{enumerate}

\section{Related Works}
\label{sec:related-works}

\textbf{Prompt tuning} shines first in natural language processing~\cite{NEURIPS2020_1457c0d6,  li2021prefixtuning, lester2021power} and then in computer vision starting from Visual Prompt Tuning (VPT)~\cite{jia2022visual}.
VPT learns parameters in the input space, so-called prompts, which are passed through the entire model together with input data.
Compared with linear probing whose performance depends on the linear separability of features~\cite{he2022masked, parashar2024neglected},
VPT leverages deep computations to non-linearly process input images towards linearly separable features used for classification.
Extending VPT, GateVPT~\cite{yoo2023improving} adaptively determines which Transformer blocks need prompt tuning.
\cite{tu2023visual, cheng2023e2vpt} append prompts only to query or key tokens in attention layers to reduce computation overhead; SPT~\cite{wang2024revisiting} re-examines the initialization of prompts and proposes to construct prompts by clustering image patch tokens.
Different from these works, we delve into the attention module, uncover burstiness in data distributions, propose several simple methods to enhance prompt tuning, and show our methods significantly outperform prior works on various benchmarks.

{\bf Burstiness} is a phenomenon first reported in natural language processing~\cite{church1995poisson, katz1996distribution}, meaning that words tend to appear in bursts, \ie, if a word appears once in a document, it is more likely to appear again. The burstiness phenomenon also exists in vision. For instance, low-level vision works report that visual element appears more times in an
image than a statistically independent model would predict~\cite{jegou2009burstiness}.
In deep learning, burstiness exists in second-order or bilinear feature learning and makes it hard to learn well~\cite{carreira2012semantic, kong2017low, lin2015bilinear}.
For the first time, we uncover burstiness in self-attention of the Transformer network architecture, where there are several matrix multiplications, \eg, multiplying query and key projectors and image tokens.

{\bf Data whitening} is a data processing technique used in machine learning that decorrelates features and equalizes their variance, transforming them to be more Gaussian.
It can remove redundancy in data and accelerate learning speed~\cite{Halkj1996the, lecun2002efficient, wiesler2011convergence}. 
Whitening is a direct inspiration for several deep learning techniques such as batch normalization~\cite{ioffe2015batch} and dynamical isometry~\cite{pennington2017resurrecting, xiao2018dynamical}.
Nowadays,
whitening is not a prerequisite and network architectures adopt normalizations such as LayerNorm~\cite{lei2016layer} and BatchNorm~\cite{ioffe2015batch}.
The main difference between them is that whitening can decorrelate features while others cannot.
In this work, after uncovering the burstiness in data involved in the attention modules in the contemporary Transformer network~\cite{vaswani2017attention, dosovitskiy2020image}, we propose whitening ${\color{Maroon}\W_q \W_k^T\X^T}$, hence more Gaussian, before learning prompt $\P$. Results show that this significantly accelerates learning and boosts accuracy.

{\bf Bilinear Model}
can model two-factor variations such as decoupling style and content in images~\cite{tenenbaum2000separating}.
It also demonstrates great success in constructing features~\cite{csurka2004visual, jegou2010aggregating, perronnin2010improving, carreira2012semantic}.
In the contemporary literature, 
bilinear deep neural networks significantly enhance visual recognition especially at a fine-grained level~\cite{lin2015bilinear, lin2017improved, kong2017low}.
Yet, it is well known that bilinear neural networks produce burstiness in the learned features, which make it hard to learn classifiers on top of them~\cite{lin2015bilinear, kong2017low}.
To address the busrtiness issue,
various normalization techniques are proposed to normalize bilinear features before feeding them to the classifier ~\cite{perronnin2010improving, koniusz2018deeper, koniusz2013comparison, lin2017improved, ionescu2015matrix}.
Different from existing bilinear models which normalize output (bilinear) features to mitigate their burstiness and facilitate learning projector weights,
ours processes input data (\eg, by whitening) and promotes learning ``bursty prompts'' with frozen projector weights of a pre-trained Transformer.

\section{Methodology: Bilinear Prompt Tuning}
\label{sec:method}

We first introduce preliminaries, then motivate a series of Bilinear Prompt Tuning (BPT) methods, and lastly provide insightful discussions and remarks.

\subsection{Preliminaries}

Let us write the conceptual formula of VPT~\cite{jia2022visual}.
Over a pretrained MODEL parameterized by $\Tha$,
VPT learns a Head ${\bf H}$ for classification and prompts $\P$ by minimizing a loss $\ell$ such as cross-entropy\footnote{For brevity, we abusively denote either an image's $n$ patch embeddings or $N$ patches from more training data as $\X$, depending on the context.}:
\begin{align}
\hat\P, \hat{\textbf{H}} = \min_{\P, \text{\bf H}} \ell \Big(\Y, \text{MODEL}(\X; \P, \text{\bf H}, \Tha)\Big),
\nonumber
\end{align}
where $\X$ and $\Y$ are the training data and ground-truth, respectively.
The MODEL has attention modules (Eq.~\ref{eq:attention1})
where the learned $\hat\P$ interacts with data through multiplications (Eq.~\ref{eq:attention3}).
As shown in Fig.~\ref{fig:splashy-figure}\textcolor{iccvblue}{b-d}, such data has burstiness and follows non-Gaussian distributions,
which intuitively make it difficult to learn $\P$.

% \begin{figure}[t]
%     \centering
%     % \subcaptionbox{Prompt gradient}
%     \includegraphics[width=0.99\linewidth]{ICCV25/figure/rebuttal_grad_v4.pdf} \\
%     \vspace{1mm}
%     % \hfill
%     % \subcaptionbox{Prompt values}
%     \includegraphics[width=0.99\linewidth]{figure/rebuttal_variable_v3.pdf} 
%     \vspace{-4mm}
% \caption{\small
%     We compare prompts' gradients $\ell_\infty$-norm (upper row) and max values (lower row) in optimization iterations by VPT and BPT respectively.
%     Results show that whitening helps BPT (1) produce more stable and larger gradients in early optimization, and (2) yield larger values in learned prompts.
%     This helps explain why BPT accelerates training.
% }
% \vspace{-3mm}
% \label{fig:learning_dynamics}
% \end{figure}

\subsection{BPT with Data Whitening}
\label{ssec:whiten}

The computations within the attention module (Eq.~\ref{eq:attention3}) have multiple parts involving $\P$, making it non-trivial to analyze how to learn $\P$ more efficiently and effectively.
We simplify our analysis by focusing on ${\color{Maroon}\P\W_q \W_k^T\X^T}$,
which involves multiplications between (supposedly)  $N$ training patch embeddings $\X \in \RB^{N\times d}$, and the key and query projectors ($\W_k$ and $\W_q$).
We denote $\tilde\X=\W_q\W_k^T\X^T \in \RB^{d \times N}$. Fig.~\ref{fig:splashy-figure}\textcolor{iccvblue}{b} shows burstiness of $\tilde\X$.
From data pre-processing perspective, we use ZCA whitening~\cite{friedman1987exploratory, hyvarinen2000independent, kessy2018optimal} to process $\tilde\X$ before learning prompts $\P$.
Denoting the covariance matrix $\SigmaBM = \frac{1}{N}\tilde\X \tilde\X^T \in \RB^{d\times d}$,
we derive a ZCA whitening matirx $\W = \SigmaBM^{-\frac{1}{2}} \in  \RB^{d\times d}$.
Concretely, we apply Singular Value Decomposition (SVD): $\SigmaBM=\U\text{\bf S}\U^T$, 
then derive the whitening matrix $\W=\SigmaBM^{-\frac{1}{2}}=\U\text{\bf S}^{-\frac{1}{2}}\U^T$.
During learning prompt $\P$, we apply the whitening matrix $\W$:
\begin{align}
\hat\P, \hat{\textbf{H}} = \min_{\P, \text{\bf H}} \ell \Big(\Y, \text{MODEL}(\X; \P\W^T, \text{\bf H}, \Tha) \Big)
\nonumber
\end{align}
In this sense, the \emph{effective learned prompt} is $\tilde\P=\hat\P\W^T$.
As the whitening matrix is computed before hand and \emph{fixed} during prompt tuning, we call the above method \textbf{BPT-fWhiten}.
From the previewed results in Table~\ref{tab:prompt-strategy}, we see that whitening makes a significant difference -- compared to the standard VPT~\cite{jia2022visual} that achieves  42.15\% and 63.71\% accuracy on the CUB-200-2011 and ImageNet benchmarks, 
BPT-fWhiten boosts to 77.48\% and 72.09\%, respectively!

Note that the MODEL is differentiable w.r.t the whitening matrix $\W$, meaning that $\W$ is \emph{tunable}.
Therefore, we present \textbf{BPT-tWhiten} which tunes both the prompt $\P$ and the whitening matrix $\W$:
\begin{align}
\hat\P, \hat\W, \hat{\textbf{H}} = \min_{\P, \W, \textbf{H}} \ell \Big(\Y, \text{MODEL}(\X; \P\W^T, \textbf{H}, \Tha) \Big)
\nonumber
\end{align}
BPT-tWhiten has the effective prompt $\tilde\P=\hat\P{\hat\W}^T$.
In Table~\ref{tab:prompt-strategy}, BPT-tWhiten outperforms BPT-fWhiten but introduces more parameters need to learn and store.
Next, we present a more compact BPT method.

\subsection{BPT with Low-Rank Bilinear Prompts}
\label{ssec:bilinear}

In BPT-tWhiten, the whitening matrix $\W$ serves as an initialization and is tuned together with prompt $\P$.
This indicates that one can learn two matrices $\A\in \RB^{m\times p}$ and $\B\in\RB^{d\times p}$, towards a \emph{bilinear} prompt $\tilde\P=\hat\A {\hat\B}^T$:
\begin{align}
\hat\A, \hat\B, \hat{\textbf{H}} = \min_{\A, \B, \text{\bf H}} \ell \Big(\Y, \text{MODEL}(\X; \A\B^T, \text{\bf H}, \Tha) \Big)
\nonumber
\end{align}
We call this BPT method {\bf BPT-bilinear} due to its two-matrix multiplication structure~\cite{lin2015bilinear, kong2017low}.
With $p < d$, we derive a low-rank prompt learning~\cite{kong2017low}.
Given the length $m$ and width $d$ of prompts, we can control $p$ to reduce the overall number of parameters and obtain a low-rank prompt $\tilde\P$.
We discuss the reduced computation in more depth in $\S$\ref{ssec:remarks}.

\subsection{Remarks}
\label{ssec:remarks}

{\bf Connections between our BPT methods.} 
BPT-fWhiten and BPT-tWhiten are motivated by data whitening, 
de-correlating the data and equalizing their variance hence
transforming them into a more Gaussian distribution.
The methods result into two-factor or bilinear models consisting of the whitening matrix and the prompt to be learned.
Following this bilinear format,
we present a compact version BPT-bilinear which learns two smaller matrices, with one resembling the whitening matrix and the other as prompt to be learned.
Hence, their computing mechanism share the same bilinear operations (\cref{fig:framework}).
Moreover, due to their bilinear operations, they learn final prompts that have burstiness as shown in \cref{fig:splashy-figure}\textcolor{iccvblue}{f-h}, and importantly, accelerate training and improve the final performance (\cref{fig:splashy-figure}\textcolor{iccvblue}{i-j}).
This suggests that, in face of burstiness data (\ie, ${\color{Maroon}\W_q \W_k^T\X^T}$), promoting learning bursty prompts eventually facilitate prompt tuning a Transformer.

{\bf Computation overhead.}
Compared with VPT~\cite{jia2022visual},
BPT-fWhiten and BPT-tWhiten introduce an extra matrix $\W$ on top of VPT \emph{during training}. 
Yet, during testing, we can multiply this matrix and the learned prompt as a single projector, which is the final effective prompt (Fig.~\ref{fig:framework}), maintaining the computation as needed in the original VPT.
Importantly, BPT-bilinear can save computation through its learned low-rank bilinear prompt~\cite{kong2017low}.
To simplify the analysis, let's focus on the data  $\tilde\X=\W_q\W_k^T\X^T \in \RB^{d \times n}$.
VPT, which learns a single prompt matrix $\P\in\RB^{m\times d}$, requires 
$m d n$ multiplications.
In comparison,
for $\A\in \RB^{m\times p}$ and $\B\in\RB^{d\times p}$,
BPT-bilinear needs 
$mpn + pdn = (mp+pd)n$
multiplications in $\A\B^T\tilde\X$.
For example, let's set the prompt length $m=100$, patch token feature dimension $d=768$, and patch length $n=196$ (commonly used in VPT~\cite{jia2022visual} and its variants~\cite{wang2024revisiting, yoo2023improving} with the 
ViT-Base/Large networks and image resolution 224$\times$224).
We set $p=25$ in BPT-bilinear (which already performs significantly better than VPT as shown in \Cref{tab:prompt-width}).
We can see that VPT and BPT need $100\times768\times196\approx 15.1\times 10^{6}$ and $(100\times25+25\times768)\times196=4.3\times 10^6$ multiplications, respectively.
That said, BPT has 3.5$\times$ computation reduction compared with VPT.

{\bf The \emph{shallow} and \emph{deep} variants.}
Like VPT~\cite{jia2022visual}, our BPT can be implemented either only in the first Transformer block (so-called \emph{shallow}) or in more blocks (so-called \emph{deep}).
Extending BPT from shallow to deep significantly improves performance (\Cref{tab:deep-variant}).
Our BPT allows controlling the compactness of learned prompt, e.g., by setting the dimensions of $\A$ and $\B$ in BPT-bilinear.
Moreover inspired by GateVPT~\cite{yoo2023improving},
we can carry out BPT-Deep in partial Transformer blocks to further reduce the number of learned parameters
while maintaining high accuracy (Table~\ref{tab:deep-variant}).

\begin{table*}[t]
\centering
\caption{
\textbf{Ablation study.}
We report top-1 accuracy (\%) on ImageNet-1K and CUB-200-2021 benchmarks and  use ViT-B whose patch token feature dimension is 768.
The number of learnable parameters (in $10^{-2}$M) excludes the task head. 
Recall that we implement our BPT using a 1$\times$1 convolution layer, where the weight is either the \emph{whitening} matrix $\W$ or \emph{random} matrix  $\B$ in BPT-bilinear.
In Table~\ref{tab:prompt-strategy}, we mark how to initialize the weight and whether to tune it. Here, the prompt length of all methods is set to 100 following VPT~\cite{jia2022visual} and SPT~\cite{wang2024revisiting}.
In Table~\ref{tab:prompt-length} and \ref{tab:prompt-width},  the default method is BPT-bilinear  (marked by {\setlength{\fboxsep}{1pt}\colorbox{col2}{red}}),
initialized with random matrix $\A\in\RB^{100\times 75}$ and $\B\in\RB^{768\times 75}$.
}
\vspace{-2.5mm}
    \begin{subtable}[t]{0.43\textwidth}
        \renewcommand{\arraystretch}{1.1}
        \centering
        \caption{Learning ``bursty'' prompts by our BPT methods outperforms prior arts VPT and SPT.
        Refer to Fig.~\ref{fig:splashy-figure} for ``bursty'' distributions.}
        \resizebox{!}{0.165\textwidth}{
        \begin{tabular}{lccrcc}
        \toprule
        methods & init. & tuning? & \#params & IN-1K & CUB-200\\
        \midrule
        \rowcolor{lightlightgrey} VPT~\cite{jia2022visual}           & -      & -            & 7.68  & 63.71 & 42.15\\
        \rowcolor{lightlightgrey} SPT~\cite{wang2024revisiting}      & -      & -            & 7.68  & 69.98 & 71.15\\
        BPT-fWhiten                         & whiten & \ding{55}    & 7.68  & 72.09 & 77.48\\
        BPT-tWhiten                         & whiten & $\checkmark$ & 66.66 & \textbf{72.37} & \textbf{78.54}\\
        \cellcolor{col2}{BPT-bilinear}  & random & $\checkmark$ & \textbf{6.51}  & 72.15 & 77.86\\
        \bottomrule
        \end{tabular}
        }
        \label{tab:prompt-strategy}
    \end{subtable}
    \hfill
    \begin{subtable}[t]{0.25\textwidth}
        \renewcommand{\arraystretch}{1.1}
        \centering
        \caption{\textbf{Prompt length}. Increasing length further improves accuracy.}
        \resizebox{!}{0.285\textwidth}{
        \begin{tabular}{cccc}
        \toprule
        length & params & IN-1K & CUB-200\\
        \midrule
        64                       & 6.24 & 71.92 & 76.91\\
        81                       & 6.37 & 71.94 & 77.60\\
        \cellcolor{col2}{100} & 6.51 & 72.15 & 77.86\\
        144                      & 6.84 & 72.39 & 78.36\\
        196                      & 7.23 & \textbf{72.48} & \textbf{78.87}\\
        \bottomrule
        \end{tabular}
        }
        \label{tab:prompt-length}
    \end{subtable}
    \hfill
    \begin{subtable}[t]{0.28\textwidth}
        \renewcommand{\arraystretch}{1.1}
        \centering
        \caption{
        \textbf{Prompt width}. BPT-bilinear learns more compact prompts with higher accuracy.
        }
        \resizebox{!}{0.255\textwidth}{
        \begin{tabular}{cccc}
        \toprule
        width & params & IN-1K & CUB-200\\
        \midrule
        \rowcolor{lightlightgrey} SPT~\cite{wang2024revisiting}      & 7.68  & 69.98 & 71.15\\
        25                      & 2.17 & 72.06 & 76.54\\
        50                      & 4.34 & 72.08 & 76.95\\
        \cellcolor{col2}{75} & 6.51 & 72.15 & 77.86\\
        100                     & 8.68 & \textbf{72.18} & \textbf{78.59}\\
        \bottomrule
        \end{tabular}
        }
        \label{tab:prompt-width}
    \end{subtable}
\vspace{-1mm}
\label{tab:ablation}
\end{table*}

\section{Experiments}
We conduct extensive experiments and ablation studies to validate our BPT methods.
First, we describe our experimental setups, including downstream tasks, datasets, pre-trained backbones, and implementation details.
Then, we carry out ablation studies and in-depth analysis on our BPT methods.
Lastly, we benchmark BPT by comparing against existing prompt tuning methods on a wide range of downstream tasks.

\subsection{Experimental Setups}
\label{sec:setup}

\textbf{Downstream tasks.}
We study BPT through the lens of three different downstream tasks: image classification, object detection and instance segmentation.

{\bf Datasets.} We use the following datasets in this work.
\begin{itemize}
    \item
    \emph{FGVC} contains five popular datasets for Fine-Grained Visual Classification, including CUB-200-2011~\cite{wah2011caltech}, NABirds~\cite{van2015building}, Oxford Flowers~\cite{nilsback2008automated}, Stanford Dogs~\cite{khosla2011novel}, and Stanford Cars~\cite{gebru2017fine}. Following VPT~\cite{jia2022visual}, we randomly split the training set into \emph{train} ($90\%$) and {\em val} ($10\%$), and use \emph{val} for hyper-parameter searching.
    \item 
    \emph{ImageNet} is a well-known classification dataset~\cite{deng2009imagenet} that consists of 1,000 classes, containing 1.28 million training images and 50k validation images.
    \item 
    \emph{COCO} 2017~\cite{lin2014microsoft} is a widely used object-level recognition benchmark consisting of 118k training images and 5k validation images, with 80 object categories. We report results on bounding box object detection (AP$^{\text{box}}$) and instance segmentation (AP$^{\text{mask}}$).
\end{itemize}

\textbf{Pre-trained backbones.} 
We follow the literature~\cite{wang2024revisiting} to use multiple backbone architectures in experiments.
We use the plain Vision Transformer (ViT-Base/Large/Huge~\cite{dosovitskiy2020image}, and a larger ViT-2B~\cite{singh2023effectiveness}) as backbones.
These backbones appear to be highly applicable in various visual and multimodal recognition tasks~\cite{dehghani2023scaling, radford2021learning}. 
We also consider different methods to pre-train these backbones, including
self-supervised pre-trained on ImageNet-1K~\cite{he2022masked, chen2021mocov3}, and supervised pre-trained on ImageNet-21K~\cite{dosovitskiy2020image}.

\begin{figure}[t]
    \centering
    \includegraphics[width=0.47\textwidth]{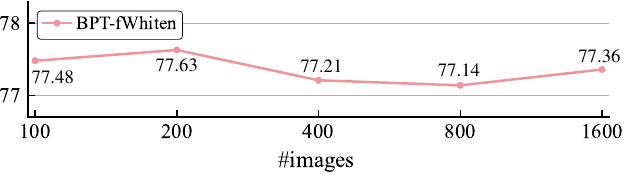}
    \vspace{-3mm}
\caption{\small
    Analysis of data scale used for calculating the whitening matrix $\W$ in BPT-fWhiten w.r.t accuracy on the CUB-200 validation set. 
    Each point is a mean accuracy over three random runs.
    Results show that it is sufficient to use a small amount of images (e.g., 100) to calculate the whitening matrix.
}
\vspace{-4.5mm}
\label{fig:whiten_patch}
\end{figure}

\textbf{Implementation details.}
We implement the two-matrix multiplication structure in BPT with a 1$\times$1 convolution~\cite{szegedy2015going, he2016deep} without bias terms,
where the weight is either the whitening matrix $\W$ or the bilinear factor $\B$.
The other matrix is the  prompt $\P$.
The 1$\times$1 layer does \emph{not} adopt any normalizations or nonlinear activations.
Other implementations follow previous works~\cite{He2019Rethinking,jia2022visual,li2022exploring,wang2024revisiting}, as described below.

For the image classification tasks, we process the image with random resize-crop to 224$\times$224 and random horizontal flip for data augmentation~\cite{jia2022visual, wang2024revisiting}. 
We use AdamW optimizer and cosine learning rate scheduler~\cite{loshchilov2016sgdr},
which gradually decays learning rate from an initial value to 1e-8.
We use linear learning rate warm-up for the 10 epochs. 

For object detection and segmentation tasks, we use the ViTDet model~\cite{li2022exploring}, in which a pre-trained ViT serves as its backbone.
We use the ViT's final feature map (16-stride, prompt tokens are discarded) to build a simple feature pyramid~\cite{li2022exploring}.
We use standard data augmentations~\cite{He2017maskrcnn, He2019Rethinking}: the image is resized with the shorter size between 480 and 800 while the longer side is no larger than 1,333.
We use AdamW optimizer and the training lasts for 3× schedule, with step-wise learning rate decay~\cite{goyal2019accurate} and linear learning rate warm-up for 1,000 iterations. The batch size is 16. 
We test using detector heads Mask R-CNN~\cite{He2017maskrcnn} and Cascade Mask R-CNN~\cite{Cai21CascadeRCNN}, respectively.

We search for the learning rate, weight decay, and epochs, for each model size (B, L, H, 2B) and for each downstream task. 
We run each method three times and report its average performance.
The hyper-parameters and more implementation details are included in the appendix.

\subsection{Ablation Study and Analysis}
\label{sec:ablation}
We perform ablation studies on the ImageNet-1K~\cite{deng2009imagenet} and CUB-200-2011~\cite{wah2011caltech} datasets
to show the effectiveness of learning ``bursty prompts'' by our BPT methods, regardless of initializations, prompt length and width, model size, training time, etc.
We report the top-1 accuracy and the number of learnable parameters\footnote{Following VPT~\cite{jia2022visual}, we  count the parameters only related to prompts, excluding those of the task head.} in \Cref{tab:ablation}. 
For ablation study, we test the shallow variant of our BPT methods (\S\ref{sec:method}), setting the prompt length to 100 and width to 75 for BPT-bilinear (the prompt width and the size of whitening matrix in BPT-fWhiten depend on the embedding dims of the ViT backbone).
We choose the self-supervised ViT-B backbone pretrained by MAE~\cite{he2022masked}. 
We discuss results below.

{\bf BPT is simple and effective.} 
\Cref{tab:prompt-strategy} compares our BPT methods and two other prompt tuning methods, VPT~\cite{jia2022visual} and SPT~\cite{wang2024revisiting}.
VPT is the very first prompt tuning method in vision, and the latter is a recent one achieving the  state-of-the-art on standard benchmarks.
All our BPT methods significantly outperform VPT and SPT.
BPT-fWhiten achieves strong gains over the SPT~\cite{wang2024revisiting}, \ie, +2.11 and +6.33 on IN-1K and CUB-200, respectively. 
Note that BPT-fWhiten learns exactly the same number of parameters as VPT and SPT. 
Recall that \cref{fig:splashy-figure}\textcolor{iccvblue}{f}  shows that BPT-fWhiten produces bursty prompts. 
The results demonstrate the effectiveness of learning bursty prompts.

By default, we randomly select 100 images from the training set to calculate the whitening matrix.
\cref{fig:whiten_patch} analyzes the affect of using different number of images to calculate the whitening matrix.
Results show that a small number of images (\eg, 100) are sufficient.
Further, when tuning the whitening matrix along with learning prompt, 
BPT-tWhiten improves by +0.3 and +1.0 points on ImageNet and CUB-200, respectively, achieving the best among these methods.
But it also introduces the most number of parameters.

Surprisingly,
BPT-bilinear learns the fewest parameters and performs on par with BPT-tWhiten.
Importantly, we randomly initialize the two matrices in BPT-bilinear. This also suggests that our BPT design is robust to random initialization, unlike SPT~\cite{wang2024revisiting} which requires carefully initializing prompts.
In the remaining of our experiments, 
we compare BPT-bilinear to prior methods.

{\bf Prompt length.}
\Cref{tab:prompt-length} studies the influence of prompt length in BPT-bilinear.
We see a positive correlation between accuracy and prompt length. 
For example, increasing prompt length from 64 to 196, BPT-bilinear improves accuracy from 71.92\% to 72.48\% on ImageNet. 
Note that previous methods (\eg, VPT~\cite{jia2022visual}, GateVPT~\cite{yoo2023improving}, and E$^2$VPT~\cite{cheng2023e2vpt}) are sensitive to prompt length (please refer to their paper).
Differently,  BPT-bilinear is easy to learn and its performance monotonically increases with larger prompt length.

{\bf Prompt width.} 
\Cref{tab:prompt-width} studies the accuracy w.r.t prompt width.
Setting a larger prompt width induces more parameters and helps BPT-bilinear perform better. Yet, the performance gains become smaller with larger width.
This demonstrates that it is sufficient to learn low-rank prompts, which demands far fewer parameters.

{
\setlength{\tabcolsep}{0.6em}
\begin{table}[t]
\centering
\caption{\small
\textbf{Analysis of BPT-Deep design choices.} 
Extending from shallow to deep, BPT achieves significant improvements.
Recall our BPT (-bilinear) allows flexible design of prompt to control the total number of learned parameters. 
We study how to design the bilinear prompt and where to insert prompt in a Transformer network.
Applying prompt tuning in all Transformer blocks achieves the best performance, but inserting prompts at the top four or six blocks is still competitive with much fewer learned parameter.
}
\vspace{-3mm}
\scalebox{0.8}{
    \begin{tabular}{l cccc}
\toprule
        variant & \#blocks & params ($\times$$10^{-2}$M) & IN-1K & CUB-200\\
\midrule
        \rowcolor{lightlightgrey}BPT-Shallow  & 1  & 6.51  & 72.15 & 77.86\\
        BPT-Deep  & 12 & 49.08 & \textbf{74.73} & \textbf{82.78}\\
        BPT-Deep  & 6 & 26.04 & 74.62 & 82.33\\
        BPT-Deep  & \cellcolor{col2}4 & 18.36 & 74.25 & 82.00\\
        BPT-Deep  & 3 & 14.52 & 74.07 & 80.86\\
    \midrule
        \rowcolor{lightlightgrey} SPT-Deep~\cite{wang2024revisiting} & 12 & 18.43 & 73.34 & 80.13\\
\bottomrule
\end{tabular}
}
\vspace{-5mm}
\label{tab:deep-variant}
\end{table}

\begin{figure*}[t]
\centering
\resizebox{1.0\textwidth}{!}{
    \subcaptionbox{scale backbone}
    {\includegraphics[width = 0.32\textwidth]{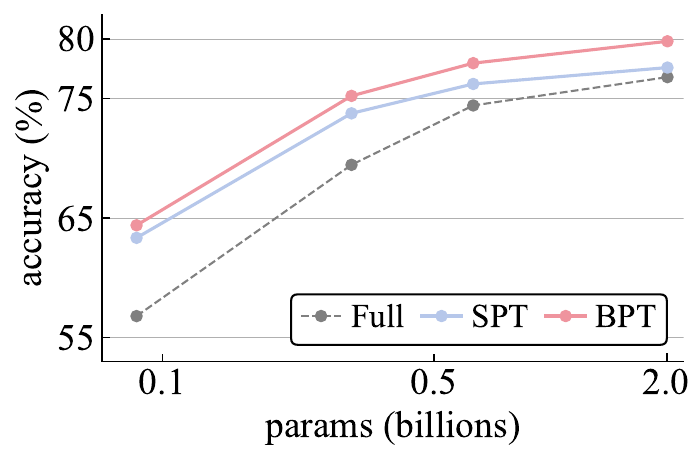}}
    \hfill
    \subcaptionbox{scale training schedule}
    {\includegraphics[width = 0.32\textwidth]{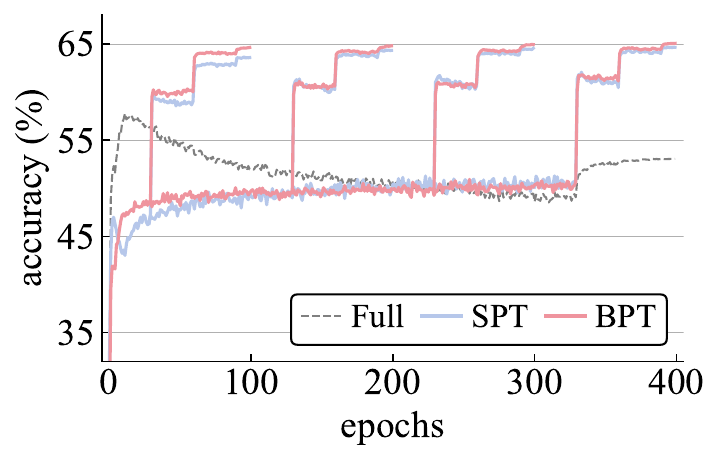}}
    \hfill
    \subcaptionbox{scale training data}
    {\includegraphics[width = 0.32\textwidth]{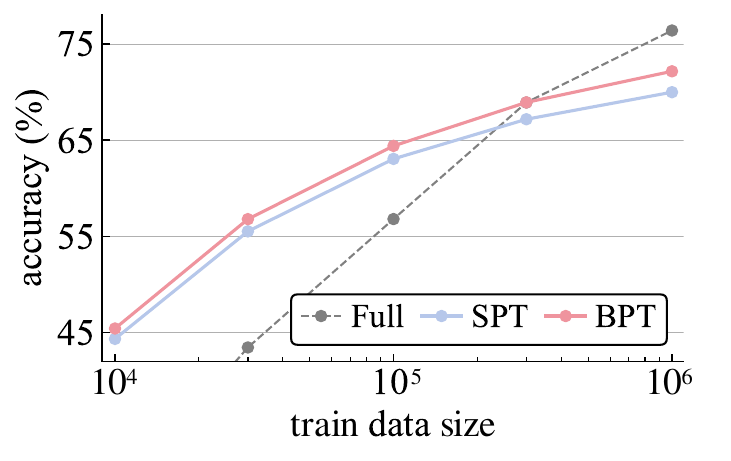}}
}
\vspace{-3mm}
\caption{\small
    \textbf{Analysis of backbone size, training epochs, and training set scale w.r.t accurarcy on ImageNet val-set.}
    All ViT backbones are self-supervised pre-trained by MAE~\cite{he2022masked}.
    Note that (a) and (b) use 10\% of the ImageNet-1K's training images.    
    Our BPT consistently outperforms the prior art SPT~\cite{wang2024revisiting} regardless of backbone size, training epochs, and training data scale.
    Yet, when training data is sufficiently large, full fine-tuning (``Full'') outshines prompt tuning methods as shown in (c).
}
\vspace{-3mm}
\label{fig:scale}
\end{figure*}
}

{\bf The deep variant of BPT} 
is derived by straightforwardly extending BPT shallow to more Transformer blocks, similar to VPT-Deep~\cite{jia2022visual} and SPT-Deep~\cite{wang2024revisiting}.
Recall that our BPT allows flexible design of prompt size to control the total number of learned parameters.
Here, we study how to design the bilinear prompt and where to insert prompt tuning in a Transformer network.
\Cref{tab:deep-variant} lists results.
First, BPT-Deep achieves significant gains over BPT-Shallow:
+2.58 on IN-1K and +4.92 on CUB-200. 
This may be due to more learned parameters in BPT-Deep,
7.54$\times$ more than BPT-Shallow.
Note that, even with a smaller prompt (50$\times$50),
inserting it to multiple Transformer blocks greatly increases the total number of learned parameters.
Then, 
we study how many top Transformer blocks to carry out prompt tuning.
The results show that,
although applying prompt tuning in all Transformer blocks achieves the best performance,
inserting prompts at the top four or six blocks lands on a middle ground w.r.t accuracy and computation.
Other works also report that fune-tuning the last few layers  achieves accuracy close to Full fine-tuning~\cite{chen2021mocov3, he2022masked, huang2023contrastive}.
Therefore, the remaining of our experiment sets the default of  BPT-Deep as learning prompts in the top four Transformer blocks.

{\bf Performance with scaled-up data, backbone size, and training epochs.}
Our study thus far is based on tuning 86M parameters with the ViT-B backbone for 100 epochs.
We now analyze the affects of using different scales of training data, different backbone sizes, and different training epochs.
We use the large-scale ImageNet to enable us to sample subsets of data for this study.
We compare our BPT-bilinear, the state-of-the-art method SPT~\cite{wang2024revisiting} and full fine-tuning in 
\cref{fig:scale},
where we use 10\% of ImageNet training data to analyze backbone size and training epochs in \cref{fig:scale}\textcolor{iccvblue}{a} and \cref{fig:scale}\textcolor{iccvblue}{b}, respectively.
We summarize salient results below.

{\em (1) Larger backbones}. 
\cref{fig:scale}\textcolor{iccvblue}{a} shows that,
increasing the backbone size from 86M to 2B,
all methods obtain a significant improvement and our BPT consistently outperforms others.
Importantly, our BPT brings more performance gains than SPT with larger backbones.
We also observe that successful full fine-tuning requires a larger drop path rate and early stopping, whereas our BPT does not require these techniques and is easier to train.

{\em (2) Longer training time}. 
% \shu{}
For convenient analysis, we use step-wise learning rate decay scheduler instead of cosine decay.
\cref{fig:scale}\textcolor{iccvblue}{b} depicts detailed results at different training epochs.
With more training epochs, full fine-tuning yields much worse results likely due to overfitting, whereas both our BPT and SPT yields better performance over time.
However, SPT also suffers from overfitting with decayed learning rates; in contrast, our BPT is quite stable and produces better results over time.
Importantly,
our BPT significantly accelerates learning and improves accuracy.
BPT within 100 epochs achieves higher accuracy than SPT within 400 epochs: 64.63\% \textit{vs}. 64.61\%.

{\em (3) More training data}. 
\cref{fig:scale}\textcolor{iccvblue}{c} shows that, with more training data, our BPT achieves larger performance gains (from +1.08 to +2.17) over the counterpart of SPT. The gains are significant in the well-studied  ImageNet benchmark. 
Moreover, BPT and SPT outperform full fine-tuning in the limited-data regime (\eg, $<$30\% of ImageNet training data), yet full fine-tune shines when more data is available for training.

{
\setlength{\tabcolsep}{0.28em}
\begin{table}[t]
\centering
\caption{\small
\textbf{Benchmarking results} with a ViT-B model self-supervised pre-trained using MAE and MoCo, 
respectively.
We compare BPT against prior methods on FGVC benchmarks. 
The best numbers for {\setlength{\fboxsep}{1.2pt}\colorbox{col1}{shallow}} (``S'') and
{\setlength{\fboxsep}{1.2pt}\colorbox{col3}{deep}} (``D'') methods are underlined and bolded, respectively.
Without bells and whistles, our BPT significantly outperforms previous methods on all the settings.
}
\vspace{-3mm}
\scalebox{0.78}{
    \begin{tabular}{l cccccc}
    \toprule
        Methods & Mean Acc & CUB & NABirds & Flowers & Dogs & Cars\\
        \midrule
        \rowcolor{lightlightgrey}\multicolumn{7}{c}{\textit{MAE pre-training}} \\
        \rowcolor{lightlightgrey} Full fine-tuning       & 82.80 & 80.55 & 77.87 & 91.71 & 80.38 & 83.51\\
        \rowcolor{col1} VPT-S~\cite{jia2022visual}  \textsubscript{ECCV'22}     & 57.84 & 42.15 & 57.43 & 69.15 & 77.07 & 43.38\\
        \rowcolor{col1} SPT-S~\cite{wang2024revisiting} \textsubscript{ICML'24} & 73.95 & 71.15 & 61.87 & 89.47 & 80.01 & 67.23\\
        \rowcolor{col1} {\bf BPT-S}  \textsubscript{ours} & \underline{80.39} & \underline{77.86} & \underline{72.03} & \underline{90.37} & \underline{81.91} & \underline{79.77}\\
        \rowcolor{col3} VPT-D~\cite{jia2022visual} \textsubscript{ECCV'22}      & 72.02 & 68.33 & 65.22 & 80.05 & 78.83 & 67.67\\
        \rowcolor{col3} GateVPT~\cite{yoo2023improving} \textsubscript{ICML'23} & 73.39 & 70.56 & 67.26 & 78.55 & 78.90 & 71.70\\
        \rowcolor{col3} SPT-D~\cite{wang2024revisiting} \textsubscript{ICML'24} & 83.26 & 80.13 & 76.28 & 93.07 & 82.23 & 84.61\\
        \rowcolor{col3} {\bf BPT-D} \textsubscript{ours} & \textbf{84.60} & \textbf{82.00} & \textbf{78.49} & \textbf{93.72} & \textbf{82.67} & \textbf{86.11}\\
        \midrule
        
        \rowcolor{lightlightgrey}\multicolumn{7}{c}{\textit{MoCo-V3 pre-training}} \\
        \rowcolor{lightlightgrey} Full fine-tuning       & 84.25 & 81.75 & 78.14 & 94.52 & 81.19 & 85.67\\
        \rowcolor{col1} VPT-S~\cite{jia2022visual}   \textsubscript{ECCV'22}    & 79.26 & 79.05 & 72.92 & 90.47 & 81.97 & 71.91\\
        \rowcolor{col1} SPT-S~\cite{wang2024revisiting} \textsubscript{ICML'24}  & 84.08 & 83.50 & 75.79 & 95.03 & 84.17 & 81.93\\
        \rowcolor{col1} {\bf BPT-S} \textsubscript{ours} & \underline{85.05} & \underline{84.39} & \underline{76.71} & \underline{95.84} & \underline{84.46} & \underline{83.84}\\
        \rowcolor{col3} VPT-D~\cite{jia2022visual} \textsubscript{ECCV'22}      & 83.12 & 82.67 & 75.99 & 94.41 & 83.33 & 79.18\\
        \rowcolor{col3} GateVPT~\cite{yoo2023improving} \textsubscript{ICML'23}  & 83.00 & 82.86 & 76.02 & 93.71 & 83.37 & 79.02\\
        \rowcolor{col3} SPT-D~\cite{wang2024revisiting} \textsubscript{ICML'24}  & 86.00 & 84.47 & 77.63 & 96.10 & 85.84 & 85.98\\
        \rowcolor{col3} {\bf BPT-D} \textsubscript{ours}  & \textbf{86.55} & \textbf{85.28} & \textbf{78.44} & \textbf{96.45} & \textbf{86.17} & \textbf{86.43}\\
        \bottomrule
    \end{tabular}
}
\vspace{-2mm}
\label{tab:sota_ssl}
\end{table}
}

\subsection{Benchmarking Results}

We now benchmark our BPT on popular datasets.
We compare our BPT-bilinear against previous methods.
For the shallow version, we set $\A\in\RB^{100\times 75}$ and $\B\in\RB^{768\times 75}$;
for the deep version, we set $\A\in\RB^{50\times 50}$ and $\B\in\RB^{768\times 50}$, and learn prompts at the top four Transformer blocks.
Moreover, we copy the results reported by the original papers to compare with the leading methods.

{\bf Comparisons under self-supervised pre-training.}
\Cref{tab:sota_ssl} compares our BPT and previous prompt tuning approaches on the FGVC benchmarks over the ViT-B backbone self-supervised pretrained by two methods, namely MAE~\cite{he2022masked} and MoCo-v3~\cite{chen2021mocov3}.
First, our BPT performs the best on these benchmarks, regardless of using MAE or MoCo-v3 pretraining.
Concretely, 
with MAE, our BPT
obtains 80.39\% mean accuracy in  the shallow setting, which learns prompt only in the very first Transformer block.
In this setting, the previous best result is 73.95\% achieved by SPT~\cite{wang2024revisiting}.
This means a significant performance gain (+6.44) by our BPT.
Moreover,
all methods perform better when using the MoCo-v3 pretraining, compared to using the MAE pretraining.
Yet, despite better performance, 
the gaps between these methods become smaller.
These results suggest that, 
while using a more generalizable, discriminative backbone can compensate weaker prompt tuning methods,
better prompt tuning can still add further performance gains.

{\bf Comparisons under supervised pre-training.} 
\Cref{tab:sota_sup} compares different PEFT methods with supervised pretrained backbone. 
Our BPT-Deep outperforms previous methods and its learned parameters are among the fewest.
The results re-confirm the effectiveness of our BPT.

\textbf{Object detection and segmentation.}
On COCO dataset, we compares linear probing, VPT~\cite{jia2022visual}, SPT~\cite{wang2024revisiting}, and our BPT (-bilinear).
Note that VPT and SPT have not been evaluated through the lens of object detection (till now!). As these methods and ours  differ only in the prompt design and do not change the backbone and detector, we implement VPT and SPT for object detection using their released open-source code.
We use the MAE~\cite{he2022masked} pre-trained ViT-B backbone with Mask R-CNN~\cite{He2017maskrcnn} and Cascade Mask R-CNN~\cite{Cai21CascadeRCNN} architectures.
\Cref{tab:det} lists their results.
First, our results show that all the prompt tuning methods outperform linear probing, demonstrate their effectiveness in more vision tasks.
Moreover, our BPT outperforms other prompt tuning methods, regardless of shallow or deep variants, on Mask R-CNN or Cascade Mask R-CNN architectures.
Example results are visualized in the appendix.

{
\setlength{\tabcolsep}{0.34em}
\begin{table}[t]
\centering
\caption{\small
\textbf{Prompt tuning (pt) \textit{vs}. backbone fune-tuning (ft)} with a ViT-B  supervised pretrained on ImageNet-21K.
For each method, we report its parameters (in $10^{-2}$M) and the mean accuracy over three random runs on five FGVC datasets.
Our BPT outperforms the compared pt methods, ft methods, and even the full fine-tuning approach, with far fewer learned parameters.
}
\vspace{-3mm}
\scalebox{0.8}{
    \begin{tabular}{l cclrc}
    \toprule
        Methods & pt & ft & \textcolor{gray}{venue\&year} & \#param. &  mean Acc (\%)\\
        \midrule
        \rowcolor{lightlightgrey} Full fine-tuning       & ~ & $\checkmark$ & \textcolor{gray}{---} & 8.60$\times 10^9$  & 88.54\\
        \rowcolor{lightlightgrey} Linear probing                      & ~ & ~            & \textcolor{gray}{---} & 0.0   & 79.32\\
        \rowcolor{lightlightgrey} Bias~\cite{zhai2019large}           & ~ & $\checkmark$ & \textcolor{gray}{arXiv 2019} & 10.71 & 88.41\\
        \rowcolor{lightlightgrey} Adapter~\cite{houlsby2019parameter} & ~ & $\checkmark$ & \textcolor{gray}{ICML 2019} & 23.70 & 85.66\\
        \rowcolor{lightlightgrey} MP~\cite{gao2023tuning}             & ~ & $\checkmark$ & \textcolor{gray}{ICCV 2023} & 112.68 & 89.38\\
        \rowcolor{lightlightgrey} LoRA~\cite{hu2021lora}              & ~ & $\checkmark$ & \textcolor{gray}{ICLR 2022} & 39.46 & 89.85\\
        \rowcolor{lightlightgrey} SSF~\cite{lian2022scaling}          & ~ & $\checkmark$ & \textcolor{gray}{NeurIPS 2022}  & 21.69 & 90.72\\
        \rowcolor{lightlightgrey} SNF~\cite{wang2023adapting}         & ~ & $\checkmark$ & \textcolor{gray}{CVPR 2023}  & 7.68  & 90.74\\
        \rowcolor{col1} VPT-Shallow~\cite{jia2022visual}        & $\checkmark$ & & \textcolor{gray}{ECCV 2022}  & 7.68 & 84.62\\
        \rowcolor{col1} SPT-Shallow~\cite{wang2024revisiting}   & $\checkmark$ & & \textcolor{gray}{ICML 2024}  & 7.68 & 90.10\\
        \rowcolor{col1} {\bf BPT-Shallow}    & $\checkmark$ & & \textcolor{gray}{\bf ours} & \underline{6.51} & \underline{90.54}\\
        \rowcolor{col3} VPT-Deep~\cite{jia2022visual}            & $\checkmark$ & & \textcolor{gray}{ECCV 2022}     & 66.98 & 89.11\\
        \rowcolor{col3} E$^2$VPT~\cite{cheng2023e2vpt}           & $\checkmark$ & & \textcolor{gray}{ICCV 2023}     & 37.98 & 89.22\\
        \rowcolor{col3} SPT-Deep~\cite{wang2024revisiting}       & $\checkmark$ & & \textcolor{gray}{ICML 2024}     & 18.43 & 91.40\\
        \rowcolor{col3} {\bf BPT-Deep}                    & $\checkmark$ & & \textcolor{gray}{\bf ours}    & \textbf{18.36} & \textbf{91.72}\\
    \bottomrule
    \end{tabular}
}
\vspace{-3mm}
\label{tab:sota_sup}
\end{table}
}

\section{Discussions}
{\bf Societal impacts.}
Datasets in our experiments are mostly balanced in per-class training data, and the adopted pretrained models may inherit biases from the pretraining data~\cite{parashar2024neglected, liu2025few}.
Our work does not sufficiently evaluate our BPT and prior methods
on imbalanced data distributions, nor investigate whether and how they are biased as well.
Future work should investigate these aspects to address potentially negative societal impacts.

{\bf Limitations and future work.}
First,
our work may oversimplify the analysis as we only focus on ${\color{Maroon}\P\W_q \W_k^T\X^T}$ in Eq.~\ref{eq:attention3}.
Future work should consider more terms therein for in-depth analysis.
Second, 
we motivate our BPT methods from the empirical observation of burstiness; yet, we feel obliged to delve into it further in the future to offer theoretical analysis, especially on its learning dynamics.
Third, we examine burstiness through the lens of prompt learning;
burstiness may exist in the entire Transformer network.
We are intrigued to study burstiness throughout the Transformer architecture, discover and mitigate potential issues.

{
\setlength{\tabcolsep}{0.37em}
\begin{table}[t]
\small
\centering
\caption{\small
\textbf{Benchmarking results} of object detection and instance segmentation on COCO dataset. We use ViT-B as the backbone self-supervised pretrained by MAE. We test Mask R-CNN and Cascade Mask R-CNN architectures, respectively.
Although not tested on these tasks,
prompt tuning outperforms the simple linear probing method.
Importantly, our BPT significantly outperforms other prompt tuning methods.
}
\vspace{-3mm}
\scalebox{0.88}{
    \begin{tabular}{lcccccc}
    \toprule
        Methods & AP$^{\text{box}}$ & AP$^{\text{box}}_{\text{75}}$ & AP$^{\text{box}}_\text{s}$ & AP$^{\text{mask}}$ & AP$^{\text{mask}}_{\text{75}}$ & AP$^{\text{mask}}_\text{s}$\\
        \midrule
        \rowcolor{lightlightgrey} \multicolumn{7}{c}{\textit{Mask R-CNN}} \\
        \rowcolor{lightlightgrey} Linear probing         & 30.70 & 32.44 & 20.03 & 28.73 & 29.83 & 14.69\\
        \rowcolor{col1} VPT-Shallow~\cite{jia2022visual} & 33.98 & 36.45 & 20.94 & 31.68 & 33.21 & 14.89\\
        \rowcolor{col1} SPT-Shallow~\cite{wang2024revisiting} & 36.46 & 39.46 & 22.43 & 33.70 & 34.21 & 16.24\\
        \rowcolor{col1} \textbf{BPT-Shallow (ours)}   & \underline{38.55} & \underline{41.94} & \underline{23.93} & \underline{35.35} & \underline{37.43} & \underline{17.42}\\
        \rowcolor{col3} VPT-Deep~\cite{jia2022visual}  & 34.62 & 36.82 & 22.37 & 32.63 & 34.41 & 16.77\\
        \rowcolor{col3} SPT-Deep~\cite{wang2024revisiting}  & 37.85 & 41.42 & 23.51 
        & 34.71 & 36.86 & 17.22\\
        \rowcolor{col3} \textbf{BPT-Deep (ours)}       & \textbf{39.67} & \textbf{43.26} & \textbf{24.52} & \textbf{36.42} & \textbf{38.92} & \textbf{17.87}\\
        \midrule
        \rowcolor{lightlightgrey} \multicolumn{7}{c}{\textit{Cascade Mask R-CNN}} \\
        % Full fine-tuning      \\
        \rowcolor{lightlightgrey} Linear probing   & 35.12 & 38.04 & 21.81 & 31.29 & 33.08 & 16.02\\
        \rowcolor{col1} VPT-Shallow~\cite{jia2022visual}  & 36.78 & 39.63 & 21.10 & 32.63 & 34.99 & 14.93\\
        \rowcolor{col1} SPT-Shallow~\cite{wang2024revisiting}     & 39.63 & 42.93 & 24.86 & 36.78 & 39.55 & 19.18\\
        \rowcolor{col1} \textbf{BPT-Shallow (ours)} & \underline{43.01} & \underline{46.78} & \underline{26.94} & \underline{38.83} & \underline{41.95} & \underline{20.70}\\
        \rowcolor{col3} VPT-Deep~\cite{jia2022visual}             & 38.70 & 42.18 & 22.97 & 34.27 & 36.79 & 16.91\\
        \rowcolor{col3} SPT-Deep~\cite{wang2024revisiting}        & 41.32 & 44.80 & 25.53 & 37.53 & 41.54 & 20.53\\
        \rowcolor{col3} \textbf{BPT-Deep (ours)} & \textbf{44.97} & \textbf{49.42} & \textbf{28.10} & \textbf{39.69} & \textbf{43.17} & \textbf{21.36}\\
    \bottomrule
    \end{tabular}
}
\vspace{-5mm}
\label{tab:det}
\end{table}
}

\section{Conclusion}
We delve into Visual Prompt Tuning (VPT), learning a small prompt in the input space to adapt a vision Transformer (ViT).
We uncover ``burstiness'' in the ViT's attention module involving the learned prompt.
We further discover Laplacian and hyper-Laplacian distributions of entries in the key and query projectors, and the patch embeddings.
Intuitively, the burstiness and non-Gaussian distributions pose challenges in prompt tuning.
To tackle these issues, we start by whitening data towards a more Gaussian distribution.
We multiply the whitening matrix with the prompt to be learned, resulting into a bilinear format of prompt.
Surprisingly, this method significantly boosts accuracy and accelerates prompt tuning;
interestingly, it also learns ``bursty'' prompts.
Inspired by the bilinear format and the burstiness observation (as well-known in bilinear models),
we introduce several Bilinear Prompt Tuning (BPT) methods.
Extensive experiments demonstrate that BPT achieves state-of-the-art performance on various benchmark datasets.

\section*{Acknowledgements}

The authors thank the anonymous reviewers for their valuable comments and suggestions.
This work was supported by Science and Technology Development Fund of Macau SAR (0067/2024/ITP2), University of Macau (SRG2023-00044-FST), the Key R\&D Program of Zhejiang (2024SSYS0012), and the Institute of Collaborative Innovation.

{
    \small
    \bibliographystyle{ieeenat_fullname}
    \bibliography{main}
}

\newpage
\clearpage
% \setcounter{page}{1}
% \maketitlesupplementary

% \section*{}
% \begin{center}
%     \emph{\bf \large Appendix}
% \end{center}

\renewcommand{\thesection}{\Alph{section}}
\renewcommand{\theHsection}{\Alph{section}}
\setcounter{section}{0} 

% \section*{}
\begin{center}
    \emph{\bf \large Appendix}
\end{center}

{\it
\hspace{1.2em} As elaborated in the main paper, we uncover a ``burstiness'' phenomenon and non-Gaussian distributions in the values resulting from the interaction  of the key projector, query projector, and image patch embeddings within the Transformer's self-attention module. 
We address these issues with several proposed methods called Bilinear Prompt Tuning (BPT).
% are motivated to learn ``bursty prompts'' (BPT)  and propose several BPT methods which either carry out data whitening prior to prompt tuning or jointly learn two matrices as ``bilinear prompts'' or regularize the singular values of prompts during training.
Experiments demonstrate that all our BPT methods significantly accelerate learning, reduce parameter count and computation, and importantly achieves the state-of-the-art over various benchmark datasets 
% % better performance than previous methods.
% Our BPT consistently outperforms existing prompt learning approaches 
across a range of model scales, dataset sizes, and pre-training objectives. We provide more implementation details and additional results in the supplemental document.
}

\section{Implementation Details}

\textbf{Datasets.}
We follow the practice of VPT~\cite{jia2022visual} to perform the split of train/val/test for 5 FGVC datasets. 
\Cref{tab:appendix_data_info} summarizes the details of the evaluated datasets used in the paper. Moreover, we randomly sample 10\% data of each category in the training set from ImageNet~\cite{deng2009imagenet} to study the affects of different training data size.

\textbf{ViT architectures.}
We use the standard ViT~\cite{dosovitskiy2020image} architectures that have a stack of Transformer blocks~\cite{vaswani2017attention}. Each block consists of a multi-head self-attention layer and an MLP layer with LayerNorm~\cite{lei2016layer}. 
Refer to \Cref{tab:appendix_vit} for details about the models.

\textbf{Whitening matrix $\W$ and bilinear factor $\B$} 
are  implemented using a 1$\times$1 convolution layer.
% can be implemented through a weight layer, \eg, a linear layer or a 1$\times$1 convolution layer. By default, we use 1$\times$1 convolution layer, as these two implementations have similar accuracy.
We \emph{do not} use normalizations in-between or after their multiplication of the learned prompt $\P$.
We tested applying normalizations but this decreases accuracy.
% as it will destroy the distribution of final prompts and reduce accuracy.

\textbf{MAE pre-training} does not use [CLS] token~\cite{he2022masked}. We follow the original designs and treat global average pooling on the sequence of $[\P;\X] \in \RB^{(n+m)\times d}$ as input for the classification head. We observe that using $[\P;\X]$, $\P$ or $\X$ yields similar accuracy.

\textbf{Object detection and instance segmentation.}
For object detection and segmentation tasks, we follow the influential ViTDet~\cite{li2022exploring}, which uses pre-trained ViT as backbone.
We use the ViT's final feature map (16-stride, prompt tokens are discarded) to build a simple feature pyramid~\cite{li2022exploring}.
We remove the window attention modules~\cite{vaswani2017attention, liu2021Swin} as the backbone is frozen during prompt tuning, which allows the object detector to be directly adapted high-resolution input images without concerning about reaching memory limits or slowing down training speed.
We use two hidden convolution layers for Region Proposal Networks~\cite{ren2016faster} and 4 hidden convolution layers for the RoI heads as per~\cite{wu2018group, li2022exploring}. 
These hidden convolution layers are followed by LayerNorm~\cite{lei2016layer}. The training last for 3× schedule.

\begin{table}[t]
\centering
\caption{
    Specifications of the downstream-task datasets.
    We follow the practice of VPT~\cite{jia2022visual} to split train/val/test for the five FGVC datasets.
    In addition, we also study general image classification, object detection, and instance segmentation tasks on the popular ImageNet and COCO datasets.
}
\vspace{-3mm}
\renewcommand{\arraystretch}{1.1}
\resizebox{0.49\textwidth}{!}{%
    \begin{tabular}{l rrrrr}
    \toprule
        Datasets &  Description & \# Classes & Train & Val & Test\\        
        \midrule
        \multicolumn{6}{c}{\textit{Fine-grained visual recognition tasks (FGVC)}} \\
        CUB-200~\cite{wah2011caltech} & bird classification & 200 & 5,394
        & 600 & 5,794\\
        NABirds~\cite{van2015building} & bird classification & 555 & 21,536 & 2,393 & 24,633\\
        Flowers~\cite{nilsback2008automated} & flower classification & 102 & 1,020 & 1,020 & 6,149\\
        Dogs~\cite{khosla2011novel} & dog classification & 120 & 10,800 & 1,200 & 8,580\\
        Cars~\cite{gebru2017fine} &  car classification & 196 & 7,329 & 815 & 8,041\\
        \midrule
        ImageNet~\cite{deng2009imagenet} & general classification & 1,000 & 1,281,167 & 50,000 & -\\
        COCO~\cite{lin2014microsoft} & object det. and seg. & 80 & 118,287 & 5,000 & -\\
        \bottomrule
    \end{tabular}
}
\vspace{-2mm}
\label{tab:appendix_data_info}
\end{table}

\begin{table}[t]
\centering
\caption{Model architectures.}
\vspace{-3mm}
\renewcommand{\arraystretch}{1.1}
\resizebox{0.49\textwidth}{!}{%
    \begin{tabular}{l ccrrcr}
    \toprule
        arch. &  Layers & Patch size & Embed dim & MLP size & Heads & Params\\
        \midrule
        ViT-Base  & 12 & 16 & 768  & 3,072  & 12 & 86M\\
        ViT-Large & 24 & 16 & 1,024 & 4,096  & 16 & 307M\\
        ViT-Huge  & 32 & 14 & 1,280 & 5,120  & 16 & 632M\\
        ViT-2B    & 24 & 14 & 2,560 & 10,240 & 32 & 1.89B\\
        \bottomrule
    \end{tabular}
}
\vspace{-2mm}
\label{tab:appendix_vit}
\end{table}

\begin{table}[t]
\centering
\caption{Hyper-parameters used on ImageNet and COCO.
Multiple values in a cell are for different
model sizes.
Here, \emph{lr}, \emph{wd} and \emph{dp} stand for
learning rate, weight decay, and drop path rate, respectively.
Full fine-tuning also use a layer-wise learning rate  decay.
}
\vspace{-3mm}
\renewcommand{\arraystretch}{1.1}
\resizebox{0.49\textwidth}{!}{%
    \begin{tabular}{l ccccc}
    \toprule
        Methods & batch & \textit{lr} & \textit{wd} & \textit{dp} & epochs\\
        \midrule
        \rowcolor{lightgray}\multicolumn{6}{l}{\textit{ImageNet}} \\
        Full fine tuning & 1024 & 4e-3/1e-2(2B) & 0.05 & 0.1/0.1/0.2/0.3 & 100/50/50/35\\
        BPT & 1024 & 0.1/0.2/0.2/0.3 & 0.01 & 0 & 100\\
        \midrule
        \rowcolor{lightgray}\multicolumn{6}{l}{\textit{COCO}} \\
        Full fine tuning & 16 & 1e-4 & 0.1 & 0.1 & 37\\
        BPT & 16 & 5e-4 & 0.1 & 0.0 & 37\\
        \bottomrule
    \end{tabular}
}
\vspace{-3mm}
\label{tab:appendix_hyper_params}
\end{table}

\begin{table*}[t]
\centering
\caption{Results of Fig.\ref{fig:scale}. Experiments of scale backbones and epochs use 10\% of the ImageNet-1K's training images.}
\vspace{-3mm}
\renewcommand{\arraystretch}{1.1}
\resizebox{0.95\textwidth}{!}{%
    \begin{tabular}{lccccccccccccc}
    \toprule
        ~ & \multicolumn{4}{c}{backbone} & \multicolumn{5}{c}{training data} & \multicolumn{4}{c}{training epochs}
        \\
        \cmidrule(r){2-5}  \cmidrule(r){6-10} \cmidrule(r){11-14}
        % \midrule
        Methods & ViT-B & ViT-L & ViT-H & ViT-2B & 1\% & 3\% & 10\% & 30\% & 100\% & 100 & 200 & 300 & 400\\
        \midrule
        \rowcolor{lightlightgrey}Full fine-tuning      & 56.80 & 69.46 & 74.43 & 76.82 & 27.68 & 43.46 & 56.80 & 68.91 & 76.39 & 56.80 & - & - & -\\
        % \midrule
        \rowcolor{col1} SPT-Shallow~\cite{wang2024revisiting} & 63.64 & 73.77 & 76.23 & 77.61 & 44.35 & 55.52 & 63.64 & 67.17 & 69.98 & 63.64 & 64.37 & 64.54 & 64.61\\
        \rowcolor{col1} \textbf{BPT-Shallow}  & 64.63 & 75.23 & 77.97 & 79.80 & 45.43 & 56.79 & 64.63 & 68.92 & 72.15 & 64.63 & 64.79 & 64.93 & 65.05\\
        \bottomrule
    \end{tabular}
}
\vspace{-2mm}
\label{tab:appendix_scale}
\end{table*}

% \begin{table}[t]
% \centering
% \caption{A study of which Transformer blocks used to insert prompt for the deep version of BPT. Here, ``3 + 1'' means insert prompts into the first 3 blocks and the last block.}
% \vspace{-3mm}
% \renewcommand{\arraystretch}{1.1}
% \resizebox{0.40\textwidth}{!}{%
%     \begin{tabular}{cccccc}
%     \toprule
%         4 + 0 & 3 + 1 & 2 + 2 & 1 + 3 & \cellcolor{col1}0 + 4 & interval\\
%         \midrule
%         76.83 & 78.45 & 81.66 & 81.38 & \textbf{82.00} & 80.87\\
%         \bottomrule
%     \end{tabular}
% }
% \label{tab:layer-study-for-deep}
% \vspace{-1mm}
% \end{table}

\textbf{Hyper-parameters.}
We search for the learning rate (\textit{lr}), weight decay (\textit{wd}), drop path rate (\textit{dp}), and epochs for each model size (B, L, H, 2B) in each downstream task. The hyper-parameters used for ImageNet and COCO with MAE pre-training are in \Cref{tab:appendix_hyper_params}.

% \textbf{BPT-Deep} is derived by straightforwardly extending BPT-shallow to more Transformer blocks, similar to VPT-Deep~\cite{jia2022visual} and SPT-Deep~\cite{wang2024revisiting}, and yields remarkable  performance improvements over shallow variant. However, BPT-Deep introduces more learned parameters, 7.54$\times$ more than BPT-Shallow. To reduce parameters, we study a \emph{partial prompt-tuning} protocol: only insert prompts in the \emph{last} Transformer blocks, \eg, 6 or 4. 
% This protocol was also used in other visual tuning works~\cite{yosinski2014transferable, noroozi2016unsupervised, he2022masked}. 

% We observe that the layers at which prompts are inserted have a significant impact. Table~\ref{tab:layer-study-for-deep} is a comparison and evaluated on CUB-200.
% As our default settings for deep variant, learning prompts in the last 4 blocks can achieve accuracy close to that of learning all blocks.
% This phenomenon is similar to that of partially fine-tuning deep neural networks that fine-tuning the last few layers can achieve accuracy close to Full fine-tuning~\cite{chen2021mocov3, he2022masked, huang2023contrastive}. 

% We also study an \textit{interval} sampling: we split the pre-trained backbone into 4 subsets of blocks (\eg, 3 in each subset for the 12-block ViT-B). We insert prompts in the first block of each subset. This strategy is reasonably good: it has 80.87\% accuracy, 3.0 higher than the shallow variant, but lags behind our default settings.

\section{Additional Results}

\begin{table}[t]
\centering
\caption{Per-task results on FGVC benchmarks of Table~\ref{tab:sota_sup}, with supervised pre-trained ViT-B/16 backbone.}
\vspace{-3mm}
\renewcommand{\arraystretch}{1.1}
\resizebox{0.49\textwidth}{!}{%
    \begin{tabular}{l ccccc}
    \toprule
        Methods & CUB-200 & NABirds & Flowers & Dogs & Cars\\
        \midrule
        \rowcolor{lightlightgrey} Full fine-tuning   & 87.3 & 82.7 & 98.8 & 89.4 & 84.5\\
        \rowcolor{lightlightgrey} Linear probing & 85.3 & 75.9 & 97.9 & 86.2 & 51.3\\
        % \midrule
        \rowcolor{col1} VPT-S~\cite{jia2022visual} & 86.7 & 78.8 & 98.4 & 90.7 & 68.7\\
        \rowcolor{col1} SPT-S~\cite{wang2024revisiting} & 90.2 & 85.1 & 99.5 & 89.3 & 86.4\\
        % \rowcolor{col11} \textbf{BPT-S (ours)} & 90.10 & 86.17 & 99.61 & 89.44 & 87.39\\
        \rowcolor{col1} \textbf{BPT-S (ours)} & 90.1 & 86.2 & 99.6 & 89.4 & 87.4\\
        
        \rowcolor{col3} VPT-D~\cite{jia2022visual} & 88.5 & 84.2 & 99.0 & 90.2 & 83.6\\
        \rowcolor{col3} SPT-D~\cite{wang2024revisiting} & 90.6 & 87.6 & 99.8 & 89.8 & 89.2\\
        % \rowcolor{col3} \textbf{BPT-D (ours)} & 90.51 & 88.13 & 99.87 & 90.14 & 89.94\\
        \rowcolor{col3} \textbf{BPT-D (ours)} & 90.5 & 88.1 & 99.9 & 90.1 & 89.9\\
        \bottomrule
    \end{tabular}
}
\vspace{-1mm}
\label{tab:appendix_sup}
\end{table}

\begin{figure}[t]
    \centering
    \includegraphics[width=0.99\linewidth]{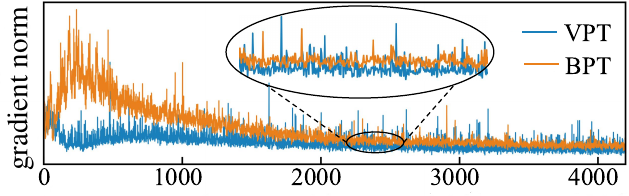} \\
    \vspace{1mm}
    \includegraphics[width=0.99\linewidth]{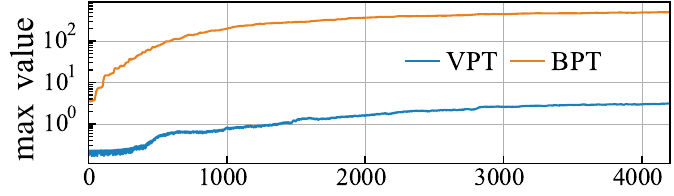} 
    \vspace{-2mm}
\caption{\small
    We compare prompts' gradients $\ell_\infty$-norm (upper row) and max values (lower row) in optimization iterations by VPT and BPT respectively.
    Results show that whitening helps BPT (1) produce more stable and larger gradients in early optimization, and (2) yield larger values in learned prompts.
    This helps explain why BPT accelerates training.
}
\vspace{-3mm}
\label{fig:learning_dynamics}
\end{figure}

\begin{figure}[t]
    \centering    
    \includegraphics[width=0.450\linewidth]{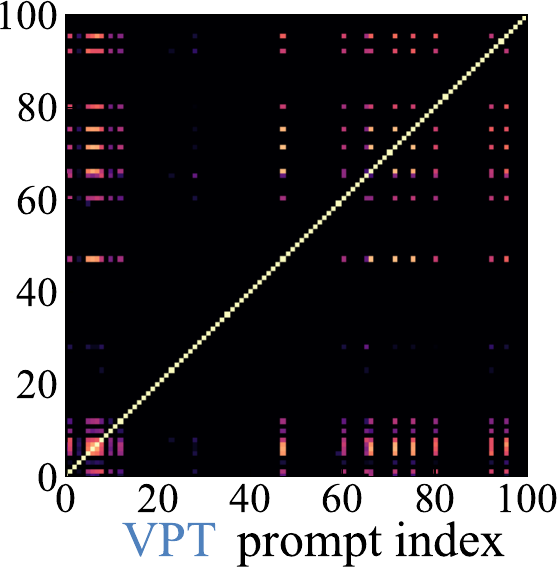} 
     \hfill
    \includegraphics[width=0.49\linewidth]{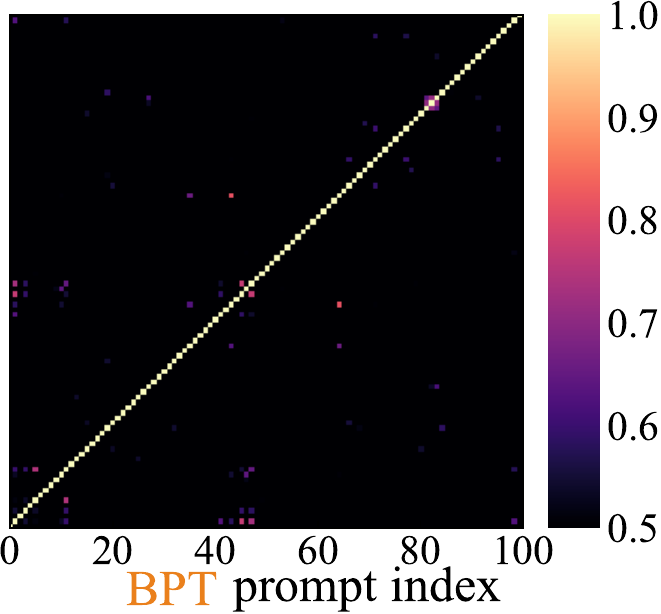}\\
    \vspace{-2mm}
\caption{\small
 Comparison of cosine similarities between learned prompts by VPT and BPT demonstrates that BPT produces more de-correlated prompts.
 This suggests BPT learns more diverse prompts than VPT.
% {\bf (D)} plots the distribution of whitened ${\color{Maroon}\W_q \W_k^T\X^T}$, non-whiten shown in Fig.~\ref{fig:splashy-figure} (b).
Zoom in for better visualization.
}
\vspace{-1mm}
\label{fig:prompt_similarity_comparison}
\end{figure}

{\bf Affects on optimization by BPT.}
As analyzing the optimization with an attention module is complex, 
we use the method in MoCo-v3~\cite{chen2021mocov3} to track prompts' gradients norm and max values in optimization iterations (Fig.~\ref{fig:learning_dynamics}). Compared with VPT~\cite{jia2022visual},
whitening helps BPT (1) produce more stable and larger gradients in early optimization, and (2) yield larger values in learned prompts.
Further, comparison of cosine similarities of learned prompts by BPT and VPT (\cref{fig:prompt_similarity_comparison}),
BPT produces more de-correlated prompts than VPT.
These analyses help explain why whitening (or BPT) accelerates learning and yields better performance than VPT.

% {\bf Comparisons between BPT and SPT on VTAB-1K.}
% We follow SPT~\cite{wang2024revisiting} and report mean accuracy on each of the three defined groups in VTAB-1K.
% VTAB-1K consists of 19 benchmarked visual tasks and is categorized into three groups. 
% Each task contains 1,000 training images. 
% We report the mean accuracy over each task group below. 
% Our BPT consistently outperforms SPT.
% \begin{table}[h]
% \centering
% % \scriptsize
% \vspace{-3.2mm}
% \renewcommand{\arraystretch}{1.1}
% \setlength\tabcolsep{10.0pt}
% \scalebox{0.750}{%
%     \begin{tabular}{lccc}
%         \toprule
%         method & Natural (7) & Specialized (4) & Structured (8)\\
%         \midrule
%         SPT-shallow~\cite{wang2024revisiting} & 62.5 & 80.9 & 53.5\\
%         BPT-shallow & {\bf 64.0} & {\bf 81.5} & {\bf 55.6}\\
%         \bottomrule
%     \end{tabular}
% }
% \vspace{-3.4mm}
% \label{tab:re-vtab}
% \end{table}

\begin{figure}[t]
\centering
    {\includegraphics[width=0.45\textwidth]{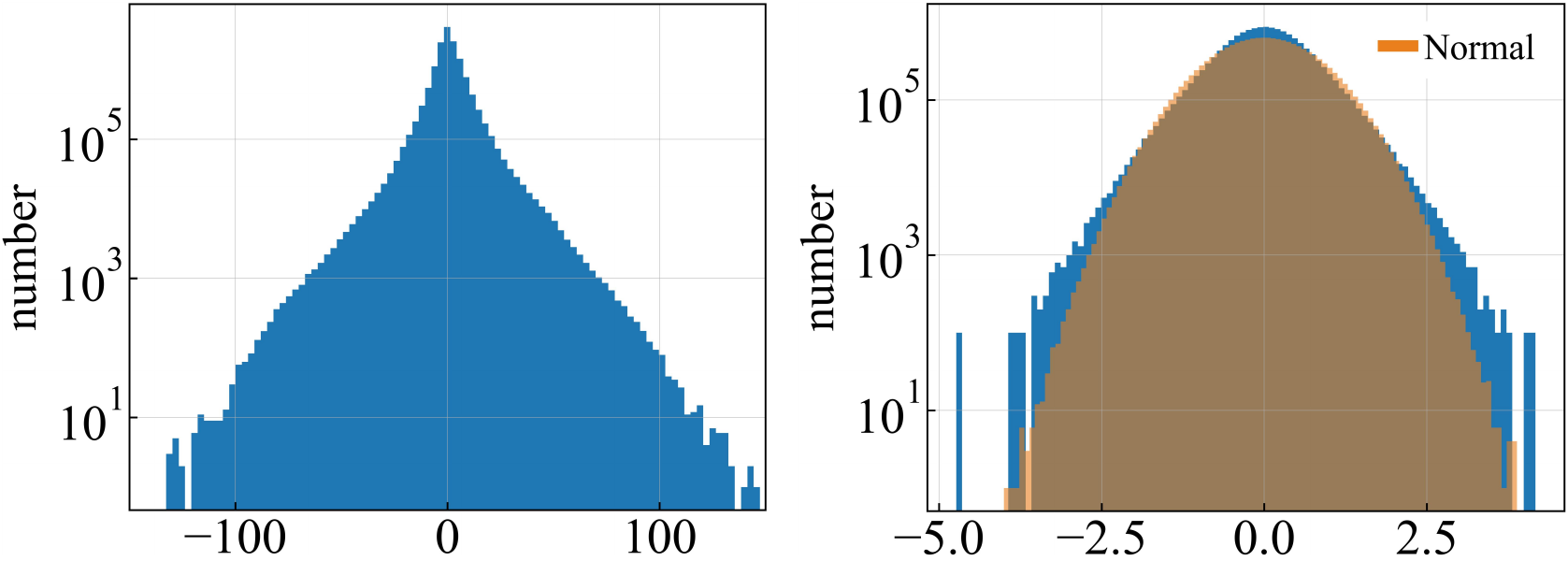}}
\vspace{-2mm}
\caption{\small
We plot the distributions of the raw (left) and whitened (right) data entries of ${\color{Maroon}\W_q \W_k^T\X^T}$.
Compared with the raw non-whitened data,
whitening (as done by BPT) transforms the data to be ``more Gaussian''.
% Zoom in for better visualization.
}
\vspace{-1mm}
\label{fig:distribution_after_whitening}
\end{figure}

{\bf Whitening transforms data to be more ``Gaussian''.}
\cref{fig:distribution_after_whitening}
plots the distribution of whitened ${\color{Maroon}\W_q \W_k^T\X^T}$.
Compared to non-whiten (left, copy from \cref{{fig:splashy-figure}}b),
whitened features follow a ``more Gaussian'' distributions.

{\bf Quantitative results of \cref{fig:scale}.}
\Cref{tab:appendix_scale} is the scale-up counterpart of Fig.~\ref{fig:scale}. All ViT backbones are self-supervised pre-trained by MAE~\cite{he2022masked} and report the top-1 accuracy on ImageNet val-set.

{\bf Per-dataset results of Table~\ref{tab:sota_sup}.}
\Cref{tab:appendix_sup} presents pre-task results on 5 FGVC datasets, with ImageNet-21K supervised pre-trained ViT-B backbone.

\begin{figure*}[t]
    \centering
    \includegraphics[width=0.98\linewidth]{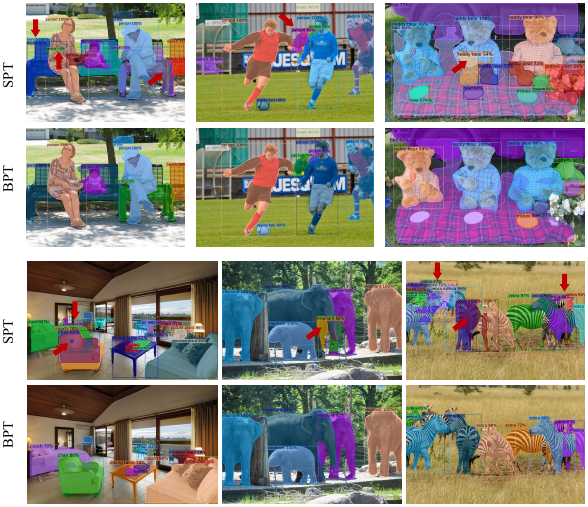}
\vspace{-2mm}
\caption{SPT~\cite{wang2024revisiting} \textit{vs}. BPT on COCO validation images. Here, MAE pre-trained ViT-B with Cascade Mask R-CNN as detector. SPT exhibits systematic artifacts on overlapping instances (marked by {\setlength{\fboxsep}{1pt}\colorbox{red}{red}} arrow).}
\vspace{-1mm}
\label{fig:appendix_det}
\end{figure*}

{\bf Visualization of detection and segmentation.}
\cref{fig:appendix_det} compares detection and segmentation results of our BPT-shallow and SPT-Shallow~\cite{wang2024revisiting} on COCO. SPT exhibits systematic artifacts on overlapping instances. Our BPT shows no such artifacts.

\begin{figure*}[t]
\centering
    \subcaptionbox{MAE pre-training}
    {\includegraphics[width = 0.99\textwidth]{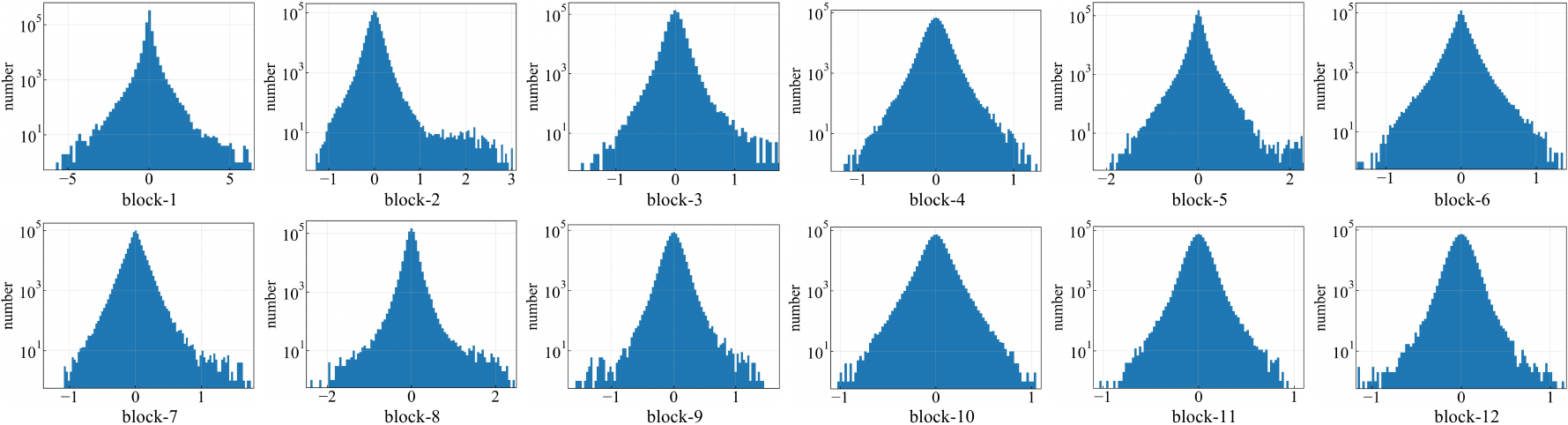}} \\
    \vspace{2mm}
    \subcaptionbox{MoCo-V3 pre-training}
    {\includegraphics[width = 0.99\textwidth]{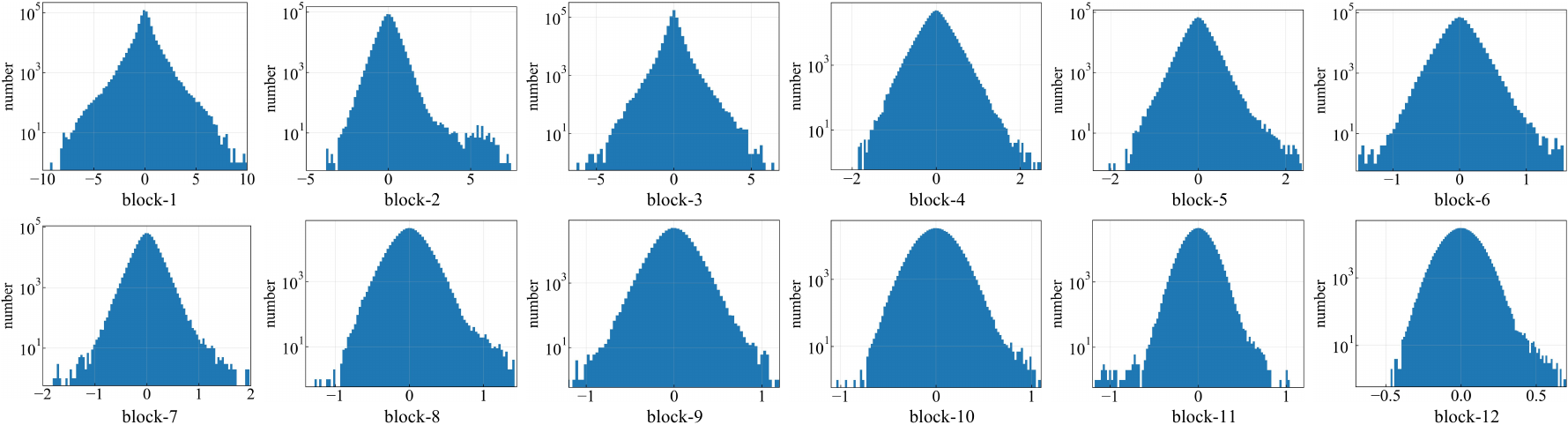}} \\
    \vspace{2mm}
    \subcaptionbox{ImageNet-21K supervised pre-training}
    {\includegraphics[width = 0.99\textwidth]{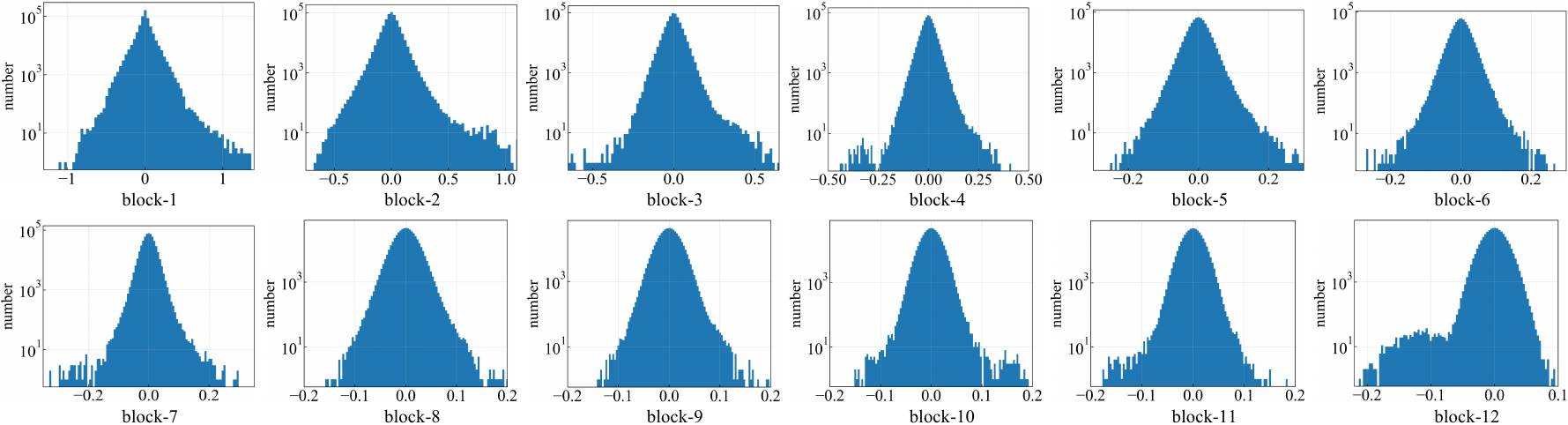}} \\
\vspace{-2mm}
\caption{
    The distribution of $\W_q \W^T_k$ w.r.t Transformer block depth.
    The backbone is ViT-B (12 blocks) and is pre-trained with three different pretraining methods (a-c).
    The three pre-training methods consistently show non-Gaussian distributions and burstiness, especially in the first blocks. 
    % For the last four blocks, MAE is more bursty than MoCo-v3 and Supervised pre-training. 
    % Supervised pre-training have smaller values.
}
\vspace{-3mm}
\label{fig:appendix_qk_distribution}
\end{figure*}

\begin{figure*}[t]
\centering
    \subcaptionbox{MAE pre-training}
    {\includegraphics[width = 0.99\textwidth]{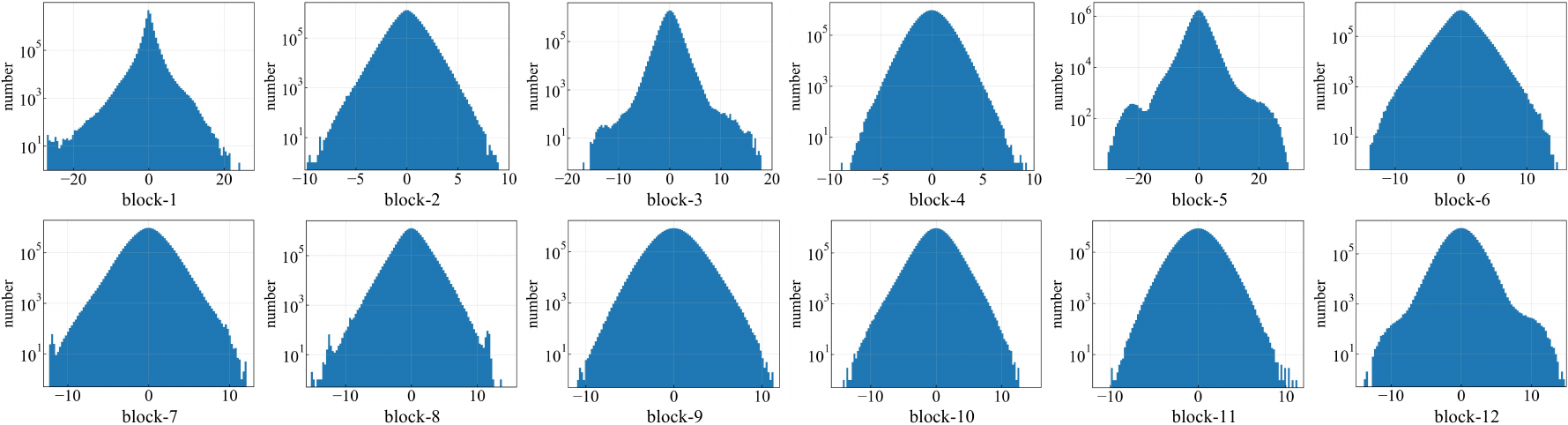}} \\
    \vspace{2mm}
    \subcaptionbox{MoCo-V3 pre-training}
    {\includegraphics[width = 0.99\textwidth]{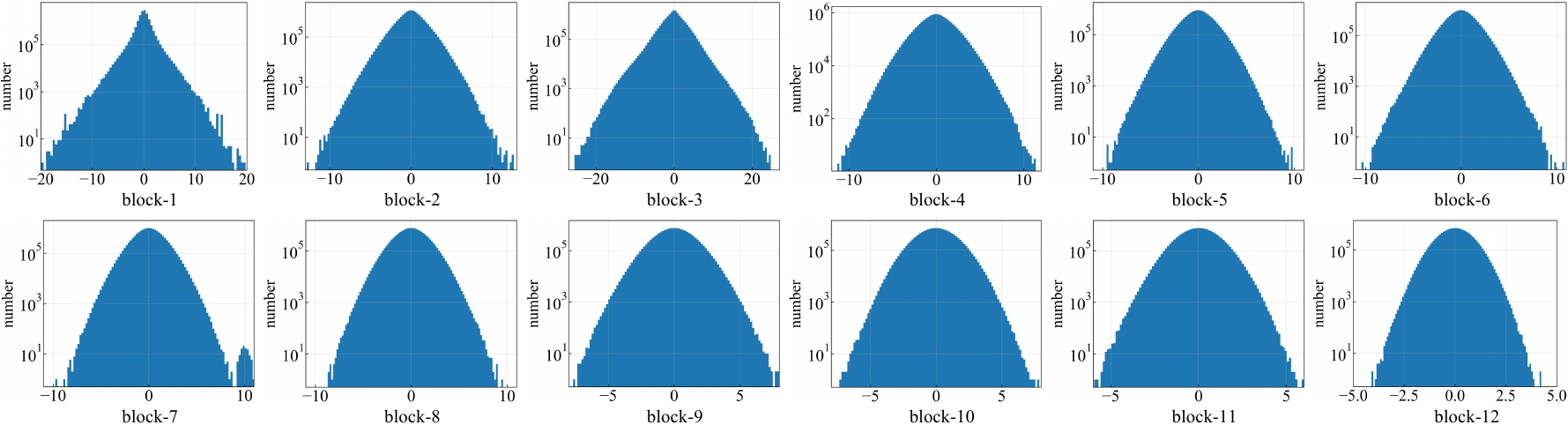}} \\
    \vspace{2mm}
    \subcaptionbox{ImageNet-21K supervised pre-training}
    {\includegraphics[width = 0.99\textwidth]{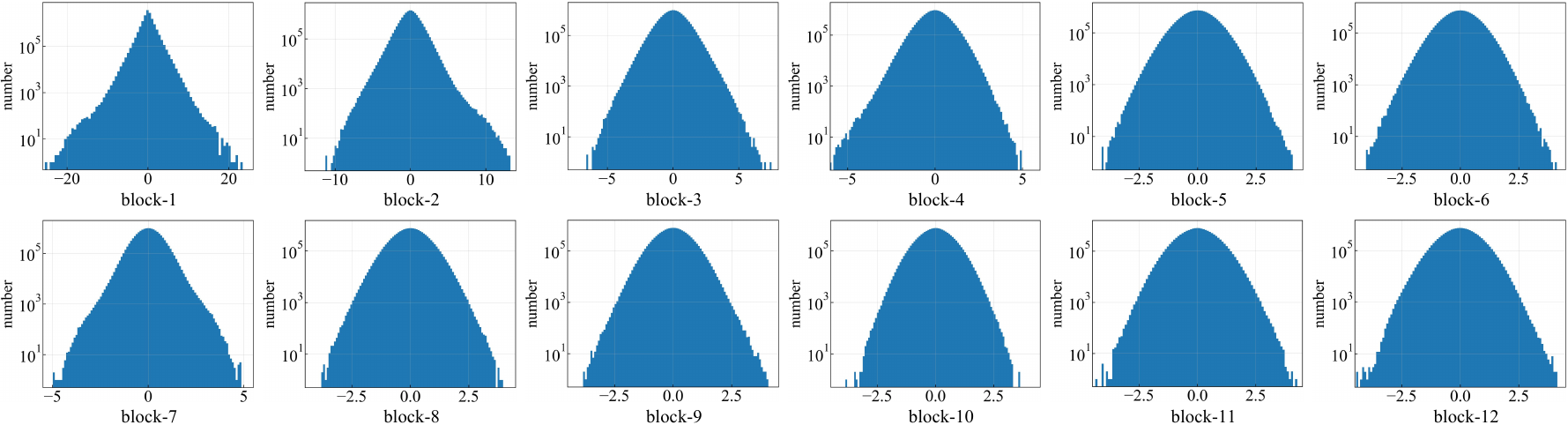}} \\
\vspace{-2mm}
\caption{
    The distribution of $\W_q \W^T_k \X^T$ w.r.t Transformer block depth.
    The backbone is ViT-B (12 blocks) and is pre-trained with three different pretraining methods (a-c).
    % The distribution of $\W_q \W^T_k \X^T$ w.r.t Transformer block depth.
    The image tokens $\X$ undergo normalizations as default implemented in typical Transformers.
    % The backbone is ViT-B which has 12 Transformer blocks.
    % and is pre-trained with 3 different objectives.
    We observe non-Gaussian distributions of these values, and the burstiness (especially in the first block regardless of training methods) which means relatively few entries have much larger values.
}
\vspace{-3mm}
\label{fig:appendix_qkx_distribution}
\end{figure*}

{\bf More distribution plots of ViT attention blocks.}
\cref{fig:appendix_qk_distribution} and \cref{fig:appendix_qkx_distribution} display distributions of entries of $\W_q \W^T_k$ and $\W_q \W^T_k \X^T$, respectively. We see non-Gaussian distributions and burstiness (especially in the first Transformer block) regardless how to pretrain the backbone (\eg, MAE, MoCO-V3, or ImageNet-21K supervised learning).

\section{Code and Demo}
\label{sec:Demo-code}

{\bf Code.}
We release our open-source codebase under the MIT License at GitHub (\url{https://github.com/WangYZ1608/BPT}.
Please refer to {\tt README.md}  for package requirements and instructions how to use the code.

{\bf Demo.}
We use Jupyter Notebook to provide demos, including plot histogram, evaluate the image classification accuracy of our BPT models and visualize the results of detection and segmentation. 
See {\tt demo-BPT-dis.ipynb}, {\tt demo-BPT-eval.ipynb}, and {\tt demo-BPT-det.ipynb} for more details at the GitHub repository.

\end{document}